\documentclass[journal]{IEEEtran}
\usepackage{epsfig}
\usepackage{subfigure}
\usepackage{graphicx}
\usepackage{amsfonts}
\usepackage{amsmath}
\usepackage{amssymb}
\usepackage{color}
\usepackage{pstricks}
\usepackage[linesnumbered,ruled]{algorithm2e}

\usepackage{dblfloatfix}
\usepackage{url}

\begin{document}
%

\title{Structured illumination microscopy image reconstruction algorithm}

%
%
%

\author{Amit~Lal, Chunyan Shan,
        and~Peng~Xi
\thanks{Amit Lal and Peng Xi are with the Department
of Biomedical Engineering, College of Engineering, Peking University, Beijing 100871, China.}%
\thanks{Chunyan Shan is with the School of Life Sciences, Peking University, Beijing 100871, China.}%
\thanks{E-mail: xipeng@pku.edu.cn}
\thanks{Manuscript received October 14, 2015.}}%

%
%

\markboth{IEEE Journal of Selected Topics in Quantum Electronics,~Vol.~X, No.~X, December~2015}{Lal \MakeLowercase{\textit{et al.}}: SIM reconstruction algorithm}
%


\IEEEspecialpapernotice{(Invited Paper)}

\maketitle

\begin{abstract}
Structured illumination microscopy (SIM) is a very important super-resolution microscopy technique, which provides high speed super-resolution with about two-fold spatial resolution enhancement. Several attempts aimed at improving the performance of SIM reconstruction algorithm have been reported. However, most of these highlight only one specific aspect of the SIM reconstruction -- such as the determination of the illumination pattern phase shift accurately -- whereas other key elements -- such as determination of modulation factor, estimation of object power spectrum, Wiener filtering frequency components with inclusion of object power spectrum information, translocating and the merging of the overlapping frequency components -- are usually glossed over superficially. In addition, most of the work reported lie scattered throughout the literature and a comprehensive review of the theoretical background is found lacking. The purpose of the present work is two-fold: 1) to collect the essential theoretical details of SIM algorithm at one place, thereby making them readily accessible to readers for the first time; and 2) to provide an open source SIM reconstruction code (named OpenSIM), which enables users to interactively vary the code parameters and study it's effect on reconstructed SIM image.

\end{abstract}


%
\IEEEpeerreviewmaketitle

\section{Introduction}
%
%
%
%
\IEEEPARstart{O}{ptical} microscopy plays critical role in life sciences study. However, the resolution of optical microscopy is limited by diffraction. The super resolution technique, owing to its ability to break this diffraction barrier has played a significant role in enabling advanced biological research, and was awarded Nobel Prize in Chemistry~2014~\cite{weiss2014nobel}. In the past decade, various super-resolution techniques aiming to break the diffraction barrier have been developed. These may be broadly categorized into three approaches~\cite{Hell2010far}: (1) point spread function (PSF) or equivalent Optical Transfer Function (OTF) modulation methods such as, stimulated emission depletion (STED)~\cite{hell1994breaking,Liu2012Achieving0,yang2014sub,chen2015subdiffraction}, reversible saturated excitation (SAX)~\cite{fujita2007high}, structured illumination microscopy (SIM)~\cite{gustafsson2000surpassing}; (2) single-molecule localization methods such as, photoactivated localization microscopy (PALM)~\cite{betzig2006imaging,hess2006ultra}, stochastic optical reconstruction microscopy (STORM)~\cite{rust2006sub}; and (3) blinking/fluctuation statistics methods such as, super-resolution optical fluctuation imaging (SOFI)~\cite{dertinger2009fast,dertinger2012superresolution,zeng2015fast}, Spatial Covariance Reconstructive (SCORE)~\cite{deng2014spatial}.

In live cell study, the temporal resolution is as important as the spatial resolution. However, current super-resolution techniques obtain ultra-high spatial resolution by  sacrificing temporal resolution~\cite{kapanidis2009biology}. As a result, it is very difficult to apply super resolution~$<$50nm to \textit{in vivo} live cell imaging. However, SIM and SOFI have demonstrated their capability in real time live cell imaging~\cite{kner2009super}.

Among the super-resolution techniques, SIM is a revolutionary subset because it brings super-resolution from the perspective of frequency domain, whereas the other super-resolution techniques, such as STED, PALM/STORM, focus on modulation in spatial domain. Moreover, SIM relaxes the requirements on sample preparation dramatically; any fluorescent sample that is used in wide-field fluorescence microscopy, \emph{is} compatible with SIM. Since its invention, SIM has been combined with different imaging modalities such as Total-internal reflection fluorescence microscopy (TIRF)~\cite{li2015extended}, surface plasmons~\cite{wei2010plasmonic}, optical section with structured illumination (HiLo)~\cite{mertz2010scanning,mertz2011optical}, light sheet microscopy~\cite{keller2010fast,chen2014lattice}, quantitative fluorescence analysis~\cite{gao2015qsim}, etc. in order to improve its resolution. It has also inspired a series of developments in super-resolution microscopy, such as saturated structured illumination microscopy (SSIM)~\cite{gustafsson2005nonlinear,rego2012nonlinear}, image scanning microscopy (ISM)~\cite{schulz2013resolution,muller2010image}, SIM with speckle pattern (blind-SIM)~\cite{mudry2012structured}, etc.

The principle of SIM OTF reconstruction was proposed by Heintzmann and Cremer in~1999~\cite{heintzmann1999laterally}. The structured illumination required can be generated by using a grating, or through digital light modulation. Due to the imperfections in the experiments, as well as the inevitable uncertainty in the measurement of experimental parameters, such as the phase and angle of illumination pattern, the SIM image reconstruction process is quite challenging. Consequently, considerable effort is devoted to development of SIM reconstruction algorithm~\cite{schaefer2004structured,shroff2010lateral,wicker2013phase,wicker2013non}. However, most of these results lie scattered in literature, which makes it very difficult for a newcomer to see the forest with a bird's-eye view. Though some open-source SIM packages exist~\cite{SIMcuda,kvrivzek2016simtoolbox,ball2015simcheck}, there lacks a general open-source SIM reconstruction \textit{code}, which hinders the development of the field, as one has to develop the entire SIM algorithm first rather than build on previous works.

Recent developments in SIM reconstruction algorithm (SIM-RA) are aimed at improving a single aspect, the accuracy of determination of the phases of sinusoidal illumination pattern~\cite{shroff2010lateral,wicker2013phase,wicker2013non}. However, there are other important aspects to SIM-RA too - for example, (1) determination of modulation factor, (2) translocating the separated frequency components to their correct locations in frequency space, (3) merging the overlapping frequency components effectively, etc. - which are usually glossed over superficially in SIM literature. This manuscript attempts to provide such information in detail.

Specifically, this manuscript is aimed at providing the bare-essential details, some of which are critical but often under-emphasized in literature, that are necessary to create a SIM code for reconstructing artifact-free super-resolved images. We first review the general SIM reconstruction principle (section~\ref{secSIMformulation}), and then present the basic SIM-RA with step by step details (section~\ref{secSIMalgorithm}). Thirdly, we present the simulated and experimental results (section~\ref{secResults}). Finally we give the discussion and conclusion. Along with this manuscript, an open-source Matlab-based SIM-RA code is also made available, so that the readers can get an in-depth understanding, and develop further algorithms based on it. This manuscript aims at connecting the ``missing links'' in SIM by providing in-depth theoretical explanation along with the SIM reconstruction source code. This enables the users with biological background to understand the parameters used for optimizing the SIM reconstruction, and to modify them so that artifacts in SIM reconstructed image is reduced. The SIM-RA presented also makes a few improvements on the algorithms reported in literature.

\section{SIM concept}
\begin{center}
\begin{figure}[t!]
\begin{pspicture}(0.0,0.1)(8.4,4.0)
\rput[lb](0.0,0.0){\scalebox{0.9}{\includegraphics[trim=167pt 659pt 164pt 66pt,clip]{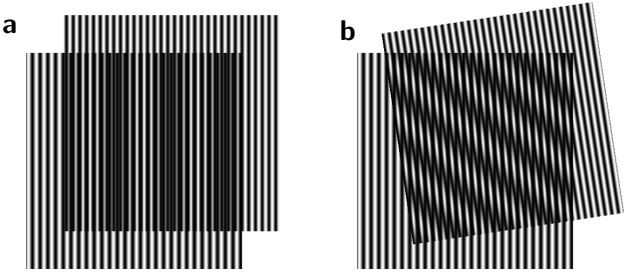}}}
\end{pspicture}
\caption{Moir\'{e} pattern formation: Superposition of two \emph{high} frequency spatial patterns results in a visually evident \textit{low} frequency spatial pattern. Figure depicts, superposition of two sinusoidal spatial patterns when their frequency vectors, in reciprocal space, are (\textsf{a}) parallel and (\textsf{b}) non-parallel.}
\label{FigMoire}
\end{figure}
\end{center}

Structured Illumination Microscopy (SIM) is a widefield super-resolution imaging technique in which a fluorescent labeled specimen is illuminated by a structured pattern of light intensity, typically sinusoidal, to effect Moir\'{e} pattern formation (Fig.~\ref{FigMoire}). By measuring the details of frequency content of the Moir\'{e} pattern in the observed image, and since the frequency content of illumination pattern is known beforehand, it is mathematically possible to compute the unknown frequency content of the specimen, theoretically up to twice the frequency limit that is conventionally imposed by the optical transfer function (OTF) of the optical system. Thus, super-resolution is achieved. Mathematical details of the technique is briefly reviewed in section~\ref{secSIMformulation}.

\begin{center}
\begin{figure*}[!t]
\begin{pspicture}(-1.0,0.0)(15,9.0)
\rput[lb](0.0,0.0){\scalebox{1.0}{\includegraphics[trim=102pt 523pt 60pt 65pt,clip]{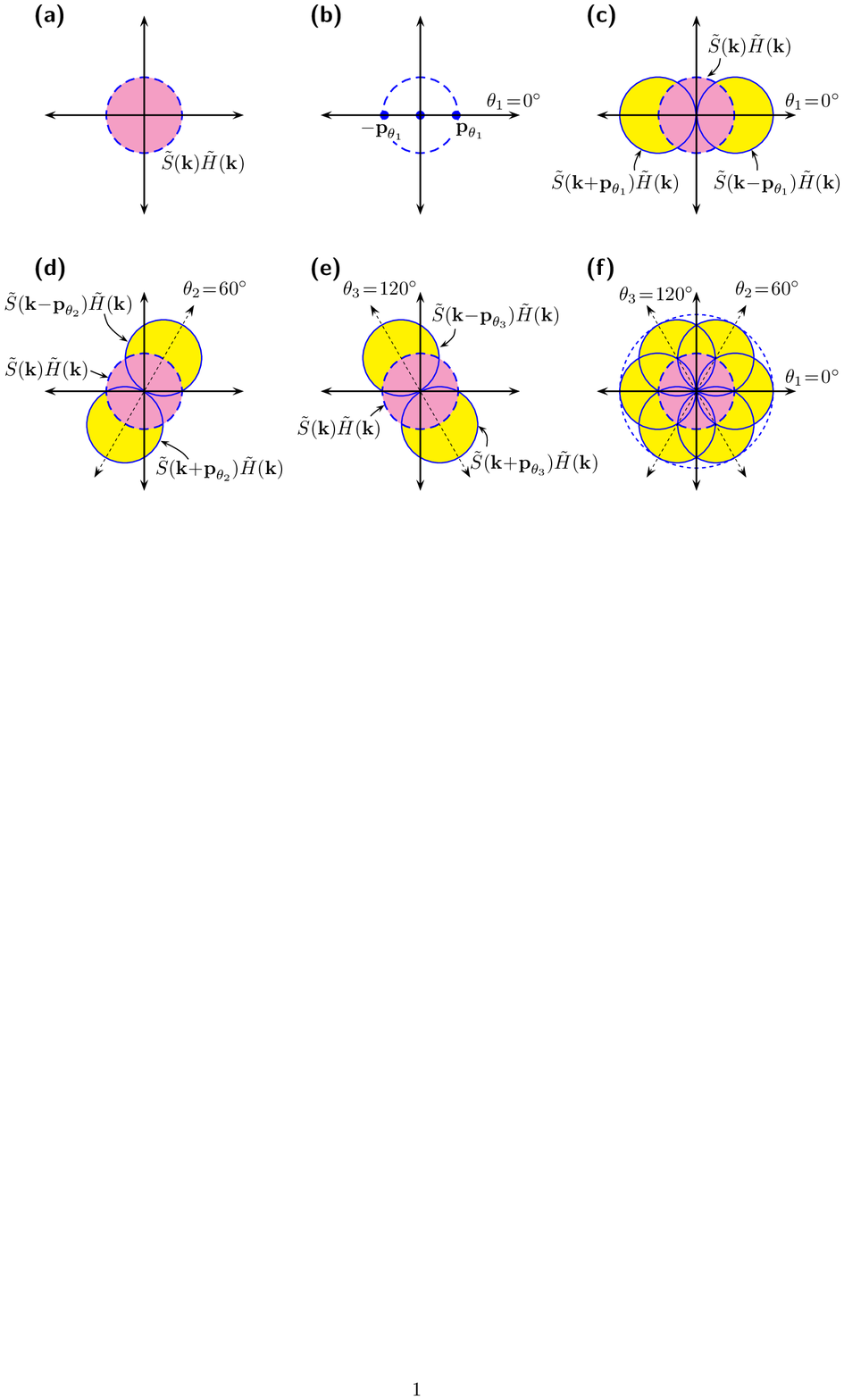}}}
\end{pspicture}
\caption{SIM concept: (a) Observable frequency content of specimen in reciprocal space is limited by optical system OTF, $\tilde{H}(\textbf{\textrm{k}})$. (b) Frequency content of sinusoidally varying intensity pattern (vertical stripes, $\theta_{1}=0^{\circ}$) relative to optical system OTF. (c,d,e) Observed frequency content of structured illuminated specimen is linear combination of frequency content within three circular regions, see Eq.~(\ref{eqSuperposed}). Note that frequency content within crescent shaped yellow regions are now observable due to Moir\'{e} effect and may be analytically computed, Eq.~(\ref{eqInvApprox}). By illuminating the specimen sequentially with sinusoidally varying illumination pattern at three different angular orientations -- say 0$^{o}$, 60$^{o}$ and 120$^{o}$ -- specimen's frequency information till \emph{twice} to that limited by optical system OTF may be obtained. (f) Separately obtained frequency contents are eventually merged and subsequently, used to construct super-resolved image of the specimen.}
\label{FigSIMconcept}
\end{figure*}
\end{center}

\vspace{-1.2cm}
\section{SIM formulation}\label{secSIMformulation}
Let $S(\textbf{\textrm{r}})$ represent fluorophore density distribution within specimen and $I_{\theta,\phi}(\textbf{\textrm{r}})$ be illuminating sinusoidal intensity pattern,
\begin{equation}
I_{\theta,\phi}(\textbf{\textrm{r}})=I_{o}\left[1-\frac{m}{2}\cos\left(2\pi \textbf{\textrm{p}}_{\theta}\cdot\textbf{\textrm{r}}+\phi\right)\right] \label{eqIllumination}
\end{equation}
where $\textbf{\textrm{r}}\equiv\left(x,y\right)$ is the (two dimensional) spatial position vector, $I_{o}$ is peak illumination intensity, $\textbf{\textrm{p}}_{\theta}=(p\cdot\cos\theta,p\cdot\sin\theta)$ is (sinusoidal) illumination frequency vector in reciprocal space, $\phi$ is phase of illumination pattern and $m$ is modulation factor. Subscript $\theta$ indicates the orientation of sinusoidal illumination pattern. Thus, the fluorescence emission distribution from specimen is $S(\textbf{\textrm{r}})\cdot I_{\theta,\phi}(\textbf{\textrm{r}})$, and the observed emission distribution through the optical system is
\begin{equation}
D_{\theta,\phi}(\textbf{\textrm{r}})=\left[S(\textbf{\textrm{r}})I_{\theta,\phi}(\textbf{\textrm{r}})\right]\otimes H(\textbf{\textrm{r}})+N(\textbf{\textrm{r}}) \label{eqImage}
\end{equation}
where $H(\textbf{\textrm{r}})$ is optical system's PSF, $\otimes$ is convolution operator and $N(\textbf{\textrm{r}})$ is additive Gaussian (white) noise.

By using convolution theorem, it may be illustrated that Fourier transform of observed image is given by (refer Supporting Information of~\cite{rego2012nonlinear})
\begin{eqnarray}
\tilde{D}_{\theta,\phi}(\textbf{k}) \hspace{-0.2cm}&=&\hspace{-0.2cm} \left[\tilde{I}_{\theta,\phi}(\textbf{\textrm{k}})\otimes\tilde{S}(\textbf{\textrm{k}})\right] \cdot \tilde{H}(\textbf{\textrm{k}}) +\tilde{N}(\textbf{\textrm{k}})  \nonumber \\
\hspace{-0.2cm}&=&\hspace{-0.2cm} \frac{I_{o}}{2}\left[\tilde{S}(\textbf{k})-\frac{m}{2}\tilde{S}(\textbf{k}-\textbf{\textrm{p}}_{\theta})e^{-i\phi}\right. \nonumber \\ &~&\hspace{0.5cm}\left.-\frac{m}{2}\tilde{S}(\textbf{k}+\textbf{\textrm{p}}_{\theta})e^{i\phi}\right] \cdot \tilde{H}(\textbf{\textrm{k}})+\tilde{N}(\textbf{\textrm{k}}) \label{eqSuperposed}
\end{eqnarray}
where $\tilde{H}(\textbf{\textrm{k}})$ is system OTF. Equation~(\ref{eqSuperposed}) suggests that $\tilde{D}_{\theta,\phi}(\textbf{k})$ is a linear combination of frequency content within three circular regions of specimen $\tilde{S}(\textbf{k})$; centered at origin, $-\textbf{\textrm{p}}_{\theta}$ and $\textbf{\textrm{p}}_{\theta}$ in reciprocal space (Fig.~\ref{FigSIMconcept}c). Consequently, three different SIM images -- $D_{\theta,\phi_{1}}(\textbf{\textrm{r}})$, $D_{\theta,\phi_{2}}(\textbf{\textrm{r}})$ and $D_{\theta,\phi_{3}}(\textbf{\textrm{r}})$ -- of the specimen is acquired corresponding to three different illumination phases; typically, $\phi_{1}=0^{\circ}$, $\phi_{2}=120^{\circ}$ and $\phi_{3}=240^{\circ}$. Then by Eq.~(\ref{eqSuperposed}), we have
\begin{equation*}
\left[ \begin{array}{c}
\tilde{D}_{\theta,\phi_{1}}(\textbf{\textrm{k}}) \\
\tilde{D}_{\theta,\phi_{2}}(\textbf{\textrm{k}}) \\
\tilde{D}_{\theta,\phi_{3}}(\textbf{\textrm{k}}) \end{array} \right]= \frac{I_{o}}{2}\textbf{\textrm{M}} \left[ \begin{array}{c}
\tilde{S}(\textbf{\textrm{k}})\tilde{H}(\textbf{\textrm{k}}) \\
\tilde{S}(\textbf{\textrm{k}}-\textbf{\textrm{p}}_{\theta})\tilde{H}(\textbf{\textrm{k}}) \\
\tilde{S}(\textbf{\textrm{k}}+\textbf{\textrm{p}}_{\theta})\tilde{H}(\textbf{\textrm{k}}) \end{array} \right] + \left[ \begin{array}{c}
\tilde{N}_{\theta,\phi_{1}}(\textbf{\textrm{k}}) \\
\tilde{N}_{\theta,\phi_{2}}(\textbf{\textrm{k}}) \\
\tilde{N}_{\theta,\phi_{3}}(\textbf{\textrm{k}}) \end{array} \right] \label{eqSuperposedMat}
\end{equation*}
\begin{equation}
\textrm{where}~~\textbf{\textrm{M}}=\left[ \begin{array}{ccc}
1 & -\frac{m}{2}e^{-i\phi_{1}} & -\frac{m}{2}e^{+i\phi_{1}} \\
1 & -\frac{m}{2}e^{-i\phi_{2}} & -\frac{m}{2}e^{+i\phi_{2}} \\
1 & -\frac{m}{2}e^{-i\phi_{3}} & -\frac{m}{2}e^{+i\phi_{3}} \end{array} \right] \label{eqImageMatrix}
\end{equation}
It may be noted from Eqs.~(\ref{eqSuperposed}) and~(\ref{eqImageMatrix}) that the constant factor $I_{o}/2$ acts trivially as to scale the intensity of captured image and thus may be assumed to be 1. Consequently, from Eq.~(\ref{eqImageMatrix}) we have
\begin{equation}
\begin{array}{r}
\textrm{noisy }\\
\textrm{estimate}\\
\textrm{of} \end{array}
\left[ \begin{array}{c}
\tilde{S}(\textbf{\textrm{k}})\tilde{H}(\textbf{\textrm{k}}) \\
\tilde{S}(\textbf{\textrm{k}}-\textbf{\textrm{p}}_{\theta})\tilde{H}(\textbf{\textrm{k}}) \\
\tilde{S}(\textbf{\textrm{k}}+\textbf{\textrm{p}}_{\theta})\tilde{H}(\textbf{\textrm{k}}) \end{array} \right] = \textbf{\textrm{M}}^{-1}\left[ \begin{array}{c}
\tilde{D}_{\theta,\phi_{1}}(\textbf{\textrm{k}}) \\
\tilde{D}_{\theta,\phi_{2}}(\textbf{\textrm{k}}) \\
\tilde{D}_{\theta,\phi_{3}}(\textbf{\textrm{k}}) \end{array} \right]
\label{eqInvApprox}
\end{equation}
Subsequently, the ungraded approximations of $\tilde{S}(\textbf{\textrm{k}})$, $\tilde{S}(\textbf{\textrm{k}}-\textbf{\textrm{p}}_{\theta})$ and $\tilde{S}(\textbf{\textrm{k}}+\textbf{\textrm{p}}_{\theta})$ are obtained by Wiener filtering of their corresponding noisy estimates obtained by above equation. Finally, the centers of frequency components $\tilde{S}(\textbf{\textrm{k}}-\textbf{\textrm{p}}_{\theta})$ and $\tilde{S}(\textbf{\textrm{k}}+\textbf{\textrm{p}}_{\theta})$ are sub-pixelly shifted to their correct locations, $+\textbf{\textrm{p}}_{\theta}$ and $-\textbf{\textrm{p}}_{\theta}$ respectively, in the reciprocal space. Thus, frequency content within crescent shaped yellow regions of Fig.~\ref{FigSIMconcept}(c), which is inaccessible by direct observation, may be computed. By changing the angular orientation $\theta$ of the illuminating sinusoidal pattern (typically, three different angular orientations -- say $\theta_{1}=0^{o}$, $\theta_{2}=60^{o}$ and $\theta_{3}=120^{o}$ -- suffices), and by repeating the above procedure, (almost) all frequency content of specimen lying within a circular region of radius twice of that governed by the OTF of optical system may be computed (Fig.~\ref{FigSIMconcept}(f)), enabling spatial reconstruction of specimen with twice the resolution than that which is directly obtainable using the same optical system. Thus, SIM reconstruction algorithm (SIM-RA) requires a set of nine different acquired images to reconstruct a super-resolved image of specimen. (Note: Though SIM reconstruction with reduced number of images~\cite{orieux2012bayesian,dong2015resolution} has been demonstrated, in the present work we restrict ourself to conventional 9-frame SIM reconstruction.)

\section{SIM Reconstruction Algorithm (SIM-RA)}\label{secSIMalgorithm}
The basic SIM-RA is presented in Algorithm~\ref{AlgoSIMcode}. The essential details necessary to carry out some of the operations of the algorithm is presented in following sections.

\begin{center}
\begin{algorithm*}[!t]
    \SetKwInOut{Input}{Input}
    \SetKwInOut{Output}{Output}

    \underline{function SIM} \;
    \Input{System OTF $\tilde{H}(\textbf{\textrm{k}})$ and nine raw SIM images $D_{\theta,\phi}(\textbf{\textrm{r}})$, corresponding to  $\theta=\theta_{1},\theta_{2},\theta_{3}$ and $\phi=\phi_{1},\phi_{2},\phi_{3}$}
    \Output{Reconstructed SIM image $D_{SIM}(\textbf{\textrm{r}})$}
    \For{$\left\{D_{\theta,\phi}(\textbf{\textrm{r}})|~\theta=\theta_{1},\theta_{2},\theta_{3}\right\}$}
    {
        \For {$\left\{D_{\theta,\phi}(\textbf{\textrm{r}})|~\phi=\phi_{1},\phi_{2},\phi_{3}\right\}$}
        {
            Estimate illumination spatial frequency $\textbf{\textrm{p}}_{\theta}$ (section~\ref{secPtheta})
        }
        Take mean of the three computed values of $\textbf{\textrm{p}}_{\theta}$ as best estimate of illumination frequency\;

        \For {$\left\{D_{\theta,\phi}(\textbf{\textrm{r}})|~\phi=\phi_{1},\phi_{2},\phi_{3}\right\}$}
        {
            Estimate illumination phase shift $\phi$ (section~\ref{secPhi})\;
            \tcc{This loop determines the three phases -- $\phi_{1},\phi_{2},\phi_{3}$}
        }

        Obtain noisy estimates of $\tilde{S}(\textbf{\textrm{k}})\tilde{H}(\textbf{\textrm{k}})$, $\tilde{S}(\textbf{\textrm{k}}-\textbf{\textrm{p}}_{\theta})\tilde{H}(\textbf{\textrm{k}})$ and $\tilde{S}(\textbf{\textrm{k}}+\textbf{\textrm{p}}_{\theta})\tilde{H}(\textbf{\textrm{k}})$ using Eq.~(\ref{eqInvApprox}) (do this by setting $m=1$ in matrix $\textbf{\textrm{M}}$ [Eq.~(\ref{eqImageMatrix})]; Adjustment for $m\neq1$ is effected later in steps~\ref{opeMf} and~\ref{opeWFcomponents} below) \; \label{opeUnmix}
    }
    Average the three noisy estimates of $\tilde{S}(\textbf{\textrm{k}})\tilde{H}(\textbf{\textrm{k}})$ (one for each $\theta=\theta_{1},\theta_{2},\theta_{3}$) and use it to estimate parameters $\mathcal{A}$ and $\alpha$ characterizing object power spectrum (section~\ref{secAAlpha}) \; \label{opeAAlpha}

    \For{$\theta=\theta_{1},\theta_{2},\theta_{3}$}
    {
        Determine modulation factor $m$ (section~\ref{secMFdetermination}) \; \label{opeMf}

        Use Wiener Filter to obtain noise-filtered and ungraded estimates $\tilde{S}_{u}(\textbf{\textrm{k}})$, $\tilde{S}_{u}(\textbf{\textrm{k}}-\textbf{\textrm{p}}_{\theta})$ and $\tilde{S}_{u}(\textbf{\textrm{k}}+\textbf{\textrm{p}}_{\theta})$ of noisy $\tilde{S}(\textbf{\textrm{k}})\tilde{H}(\textbf{\textrm{k}})$, $\tilde{S}(\textbf{\textrm{k}}-\textbf{\textrm{p}}_{\theta})\tilde{H}(\textbf{\textrm{k}})$ and $\tilde{S}(\textbf{\textrm{k}}+\textbf{\textrm{p}}_{\theta})\tilde{H}(\textbf{\textrm{k}})$, respectively (section~\ref{secWienerFilter})\; \label{opeWFcomponents}

        Shift $\tilde{S}_{u}(\textbf{\textrm{k}}-\textbf{\textrm{p}}_{\theta})$ and $\tilde{S}_{u}(\textbf{\textrm{k}}+\textbf{\textrm{p}}_{\theta})$ components to their respective true positions in frequency domain (section~\ref{secShiftFreqComp}). Let $\tilde{S}_{s}(\textbf{\textrm{k}}-\textbf{\textrm{p}}_{\theta})$ and $\tilde{S}_{s}(\textbf{\textrm{k}}+\textbf{\textrm{p}}_{\theta})$, respectively, denote components $\tilde{S}_{u}(\textbf{\textrm{k}}-\textbf{\textrm{p}}_{\theta})$ and $\tilde{S}_{u}(\textbf{\textrm{k}}+\textbf{\textrm{p}}_{\theta})$ shifted to their true positions\;

        `Phase match' the shifted components $\tilde{S}_{s}(\textbf{\textrm{k}}-\textbf{\textrm{p}}_{\theta})$ and $\tilde{S}_{s}(\textbf{\textrm{k}}+\textbf{\textrm{p}}_{\theta})$ with respect to the unshifted component $\tilde{S}_{u}(\textbf{\textrm{k}})$ (section~\ref{secShiftPhaseError}) \; \label{opeShiftCorrection}
    }
    Merge all nine frequency frequency components (three components $\tilde{S}_{u}(\textbf{\textrm{k}})$, $\tilde{S}_{s}(\textbf{\textrm{k}}-\textbf{\textrm{p}}_{\theta})$ and $\tilde{S}_{s}(\textbf{\textrm{k}}+\textbf{\textrm{p}}_{\theta})$ for each of the three $\theta s$) into one $\tilde{D}_{\textrm{SIM}}(\textbf{\textrm{k}})$ using generalized Wiener-Filter (section~\ref{secGenWF}) \; \label{opeGWFmerge}

    Compute inverse FT of $\tilde{D}_{SIM}(\textbf{\textrm{k}})$ to obtain reconstructed SIM image $D_{\textrm{SIM}}(\textbf{\textrm{r}})$ \;
    return $D_{\textrm{SIM}}(\textbf{\textrm{r}})$ \;
    \caption{SIM-RA}
    \label{AlgoSIMcode}
\end{algorithm*}
\end{center}

\vspace{-1.2cm}
\subsection{Determination of illumination spatial frequency $\textbf{\textrm{p}}_{\theta}$} \label{secPtheta}

In the SIM algorithm presented, illumination spatial frequency is determined \textit{a posteriori} by iteratively optimizing the auto-correlation of $\tilde{C}_{\theta,\phi}(\textbf{\textrm{k}})=\tilde{D}_{\theta,\phi}(\textbf{\textrm{k}}) \tilde{H}^{*}(\textbf{\textrm{k}})$ with its shifted variant $\tilde{C}_{\theta,\phi}(\textbf{k}+\textbf{\textrm{p}}_{\theta})$~:
\begin{equation}
\mathcal{C}_{1}=\sum_{\textbf{k}}\tilde{C}_{\theta,\phi}(\textbf{\textrm{k}})\tilde{C}^{*}_{\theta,\phi}(\textbf{k}+\textbf{\textrm{p}}_{\theta})
\label{eqFreqCC}
\end{equation}
Value of $\textbf{\textrm{p}}_{\theta}$ corresponding to maxima of $|\mathcal{C}_{1}|$ is the desired value of illumination spatial frequency.

\subsection{Determination of phase shift $\phi$ of illumination pattern}\label{secPhi}
Using the estimated value of illumination spatial frequency $\textbf{\textrm{p}}_{\theta}$ above, and an (arbitrary) initial guess $\phi_{o}$, a two dimensional sinusoidal function
\begin{equation}
P_{\theta,\phi_{o}}(\textbf{\textrm{r}})=-\cos\left(2\pi \textbf{\textrm{p}}_{\theta}\cdot\textbf{\textrm{r}}+\phi_{o}\right)
\label{eqPatternGuess}
\end{equation}
is constructed in the spatial domain. Using this function, an estimate of \emph{true} phase shift is obtained by iteratively optimizing the correlation
\begin{equation*}
\mathcal{C}_{2}=\sum_{\textbf{\textrm{r}}}D_{\theta,\phi}(\textbf{\textrm{r}})P_{\theta,\phi_{o}}(\textbf{\textrm{r}})
\end{equation*}
where the summation is carried over the entire range of $\textbf{\textrm{r}}$. Noting the similarity in sinusoidal parts of Eqs.~(\ref{eqIllumination}) and~(\ref{eqPatternGuess}), it is clear that $\mathcal{C}_{2}$ achieves its maxima when $\phi_{o}$ becomes $\phi$.

\subsection{Estimation of object power spectrum}\label{secAAlpha}
The noisy estimate of $\tilde{S}(\textbf{\textrm{k}})\tilde{H}(\textbf{\textrm{k}})$ that is obtained using Eq.~\ref{eqInvApprox} is used to determine the object's average power spectrum. As in~\cite{shroff2010lateral}, it is assumed here that average power spectrum of noisy $\tilde{S}(\textbf{\textrm{k}})\tilde{H}(\textbf{\textrm{k}})=|\tilde{H}(\textbf{\textrm{k}})|^{2} \mathcal{A}^{2}|\textbf{\textrm{k}}|^{-2\alpha}+\Psi_{o,\theta}$, where $\mathcal{A}$ and $\alpha$ are constants; $\Psi_{o,\theta}$ is the average noise power in $\tilde{S}(\textbf{\textrm{k}})\tilde{H}(\textbf{\textrm{k}})$ which may be estimated by averaging the square of frequency amplitude of (noisy) $\tilde{S}(\textbf{\textrm{k}})\tilde{H}(\textbf{\textrm{k}})$ over frequencies $\textbf{\textrm{k}}$ lying outside~OTF support, where signal power is zero. Consequently, an iterative non-linear regression scheme is employed to obtain the optimum values of $\mathcal{A}$ and $\alpha$, characterizing object mean power spectrum. (Reasoning for $\Psi_{o,\theta}$ being independent of $\textbf{\textrm{k}}$ is discussed in section~\ref{secWienerFilter} below.)

\subsection{Estimation of modulation factor $m$}\label{secMFdetermination}

\begin{figure*}[!t]
\normalsize
\newcounter{MYtempeqncnt}
\setcounter{MYtempeqncnt}{\value{equation}}
\begin{equation}
\textbf{\textrm{M}}^{-1}=\frac{1}{\Delta}\times\left[ \begin{array}{ccc}
e^{i\left(\phi_{2}-\phi_{3}\right)}-e^{i\left(\phi_{3}-\phi_{2}\right)} & e^{i\left(\phi_{3}-\phi_{1}\right)}-e^{i\left(\phi_{1}-\phi_{3}\right)} & e^{i\left(\phi_{2}-\phi_{1}\right)}-e^{i\left(\phi_{1}-\phi_{2}\right)} \\
\frac{2}{m}\left(e^{i\phi_{3}}-e^{i\phi_{2}}\right) & \frac{2}{m}\left(e^{i\phi_{1}}-e^{i\phi_{3}}\right) & \frac{2}{m}\left(e^{i\phi_{2}}-e^{i\phi_{1}}\right) \\
\frac{2}{m}\left(e^{-i\phi_{2}}-e^{-i\phi_{3}}\right) & \frac{2}{m}\left(e^{-i\phi_{3}}-e^{-i\phi_{1}}\right) & \frac{2}{m}\left(e^{-i\phi_{1}}-e^{-i\phi_{2}}\right)
\end{array} \right]
\label{eqnMinvLong}
\end{equation}
\begin{equation*}
\textrm{where}~~\Delta=\left[e^{i\left(\phi_{2}-\phi_{1}\right)} -e^{i\left(\phi_{1}-\phi_{2}\right)} -e^{i\left(\phi_{3}-\phi_{1}\right)} +e^{i\left(\phi_{1}-\phi_{3}\right)} +e^{i\left(\phi_{3}-\phi_{2}\right)} -e^{i\left(\phi_{2}-\phi_{3}\right)}\right]
\end{equation*}
\setcounter{MYtempeqncnt}{\value{equation}}
\setcounter{equation}{\value{MYtempeqncnt}}
\hrulefill
\vspace*{4pt}
\end{figure*}

Inverse of matrix $\textbf{\textrm{M}}$ [Eq.~(\ref{eqImageMatrix})] is given by Eq.~(\ref{eqnMinvLong}). This, in conjunction with Eq.~(\ref{eqInvApprox}), suggests that
\begin{equation}
\textrm{true~}\tilde{S}(\textbf{\textrm{k}}-\textbf{\textrm{p}}_{\theta})\tilde{H}(\textbf{\textrm{k}}) = \frac{1}{m}\left[\hspace{-0.1cm}\begin{array}{l}
\textrm{estimate of }
\tilde{S}(\textbf{\textrm{k}}-\textbf{\textrm{p}}_{\theta})\tilde{H}(\textbf{\textrm{k}}) \\
\textrm{as computed in step~\ref{opeUnmix} of} \\
\textrm{Algorithm~\ref{AlgoSIMcode} by setting~} m$=$1 \end{array}\hspace{-0.1cm}\right]
\label{eqMfacEffect}
\end{equation}
This relation is used to determine the modulation factor $m$ in SIM-RA. Using the determined values of $\mathcal{A}$ and $\alpha$ in section~\ref{secAAlpha}, average power spectrum of true $\tilde{S}(\textbf{\textrm{k}}-\textbf{\textrm{p}}_{\theta})\tilde{H}(\textbf{\textrm{k}})$ =$|\tilde{H}(\textbf{\textrm{k}})|^{2} \mathcal{A}^{2}|\textbf{\textrm{k}}-\textbf{\textrm{p}}_{\theta}|^{-2\alpha}$. Consequently, it follows that
\begin{equation}
\left.\begin{array}{r}
\textrm{Avg. power spectrum } \\
\textrm{of }\tilde{S}(\textbf{\textrm{k}}-\textbf{\textrm{p}}_{\theta})\tilde{H}(\textbf{\textrm{k}}) \\
\textrm{as computed in } \\
\textrm{step~\ref{opeUnmix} of Algorithm~\ref{AlgoSIMcode}}
\end{array} \hspace{-0.2cm}\right\}= m^{2}|\tilde{H}(\textbf{\textrm{k}})|^{2} \mathcal{A}^{2}|\textbf{\textrm{k}}-\textbf{\textrm{p}}_{\theta}|^{-2\alpha}+\Psi_{p,\theta}
\label{eqModulationFactor}
\end{equation}
where $\Psi_{p,\theta}$ is average noise power in $\tilde{S}(\textbf{\textrm{k}}-\textbf{\textrm{p}}_{\theta})\tilde{H}(\textbf{\textrm{k}})$, as computed in step~\ref{opeUnmix} of Algorithm~\ref{AlgoSIMcode}, which may be determined in a manner similar to $\Psi_{o,\theta}$, as described in the previous section. Since modulation factor $m$ is the only unknown in Eq.~(\ref{eqModulationFactor}), it may be readily determined.

\subsection{Wiener Filtering}\label{secWienerFilter}

Under the assumption that the raw SIM images are corrupted by white noise (average power spectrum of the noise is constant over all frequencies), linearity of Eqs.~(\ref{eqImageMatrix}) and~(\ref{eqInvApprox}) ensures that the estimates of $\tilde{S}(\textbf{\textrm{k}})\tilde{H}(\textbf{\textrm{k}})$, $\tilde{S}(\textbf{\textrm{k}}-\textbf{\textrm{p}}_{\theta})\tilde{H}(\textbf{\textrm{k}})$ and $\tilde{S}(\textbf{\textrm{k}}+\textbf{\textrm{p}}_{\theta})\tilde{H}(\textbf{\textrm{k}})$ determined using Eq.~(\ref{eqInvApprox}) too are corrupted by white noise. Consequently, the ungraded estimates of $\tilde{S}(\textbf{\textrm{k}})\tilde{H}(\textbf{\textrm{k}})$, $\tilde{S}(\textbf{\textrm{k}}-\textbf{\textrm{p}}_{\theta})\tilde{H}(\textbf{\textrm{k}})$ and $\tilde{S}(\textbf{\textrm{k}}+\textbf{\textrm{p}}_{\theta})\tilde{H}(\textbf{\textrm{k}})$, say $\tilde{S}_{u}(\textbf{\textrm{k}})$, $\tilde{S}_{u}(\textbf{\textrm{k}}-\textbf{\textrm{p}}_{\theta})$ and $\tilde{S}_{u}(\textbf{\textrm{k}}+\textbf{\textrm{p}}_{\theta})$, respectively, may be obtained by employing Wiener Filter~\cite{gonzalez2002woods,fienup2002comparison} as follows
\begin{equation}
\tilde{S}_{u}(\textbf{\textrm{k}}) =\left[\frac{\tilde{H}^{\ast}(\textbf{\textrm{k}})}{|\tilde{H}(\textbf{\textrm{k}})|^{2} +\frac{\Psi_{o,\theta}}{\mathcal{A}^{2}|\textbf{\textrm{k}}|^{-2\alpha}}}\right] \tilde{S}(\textbf{\textrm{k}})\tilde{H}(\textbf{\textrm{k}})
\label{eqWienerFilter1}
\end{equation}
\begin{equation}
\tilde{S}_{u}(\textbf{\textrm{k}}-\textbf{\textrm{p}}_{\theta}) =\frac{1}{m}\left[\frac{\tilde{H}^{\ast}(\textbf{\textrm{k}})}{|\tilde{H}(\textbf{\textrm{k}})|^{2} +\frac{\Psi_{p,\theta}}{m^2\mathcal{A}^{2}|\textbf{\textrm{k}}-\textbf{\textrm{p}}_{\theta}|^{-2\alpha}}}\right] \tilde{S}(\textbf{\textrm{k}}-\textbf{\textrm{p}}_{\theta})\tilde{H}(\textbf{\textrm{k}})
\label{eqWienerFilter2}
\end{equation}
\begin{equation}
\tilde{S}_{u}(\textbf{\textrm{k}}+\textbf{\textrm{p}}_{\theta}) =\frac{1}{m}\left[\frac{\tilde{H}^{\ast}(\textbf{\textrm{k}})}{|\tilde{H}(\textbf{\textrm{k}})|^{2} +\frac{\Psi_{q,\theta}}{m^2\mathcal{A}^{2}|\textbf{\textrm{k}}+\textbf{\textrm{p}}_{\theta}|^{-2\alpha}}}\right] \tilde{S}(\textbf{\textrm{k}}+\textbf{\textrm{p}}_{\theta})\tilde{H}(\textbf{\textrm{k}})
\label{eqWienerFilter3}
\end{equation}
Note that $\Psi_{o,\theta}$, $\Psi_{p,\theta}$ and $\Psi_{q,\theta}$ are average noise powers in $\tilde{S}(\textbf{\textrm{k}})\tilde{H}(\textbf{\textrm{k}})$, $\tilde{S}(\textbf{\textrm{k}}-\textbf{\textrm{p}}_{\theta})\tilde{H}(\textbf{\textrm{k}})$ and $\tilde{S}(\textbf{\textrm{k}}+\textbf{\textrm{p}}_{\theta})\tilde{H}(\textbf{\textrm{k}})$, respectively. Method of determination of $\Psi_{o,\theta}$ is described in section~\ref{secAAlpha}; $\Psi_{p,\theta}$ and $\Psi_{q,\theta}$ may be similarly determined. The extra factor $1/m$ in Eqs.~(\ref{eqWienerFilter2}) and~(\ref{eqWienerFilter3}) accounts for the fact that in step~\ref{opeUnmix} of Algorithm~\ref{AlgoSIMcode}, estimates of $\tilde{S}(\textbf{\textrm{k}}-\textbf{\textrm{p}}_{\theta})\tilde{H}(\textbf{\textrm{k}})$ and $\tilde{S}(\textbf{\textrm{k}}+\textbf{\textrm{p}}_{\theta})\tilde{H}(\textbf{\textrm{k}})$ are computed by setting $m=1$.

\subsection{Shifting frequency components $\tilde{S}_{u}(\textbf{\textrm{k}}-\textbf{\textrm{p}}_{\theta})$ and $\tilde{S}_{u}(\textbf{\textrm{k}}+\textbf{\textrm{p}}_{\theta})$ to their true positions}\label{secShiftFreqComp}

True positions of frequency components $\tilde{S}_{u}(\textbf{\textrm{k}}-\textbf{\textrm{p}}_{\theta})$ and $\tilde{S}_{u}(\textbf{\textrm{k}}+\textbf{\textrm{p}}_{\theta})$ are centered respectively at frequencies $\textbf{\textrm{p}}_{\theta}$ and $-\textbf{\textrm{p}}_{\theta}$ in the frequency domain, see Fig.~\ref{FigSIMconcept}. By employing Fourier shift theorem~\cite{gaskill1978linear}, frequency components $\tilde{S}_{u}(\textbf{\textrm{k}}-\textbf{\textrm{p}}_{\theta})$ and $\tilde{S}_{u}(\textbf{\textrm{k}}+\textbf{\textrm{p}}_{\theta})$ may be shifted to their true positions, to obtain their shifted variants (say) $\tilde{S}_{s}(\textbf{\textrm{k}}-\textbf{\textrm{p}}_{\theta})$ and $\tilde{S}_{s}(\textbf{\textrm{k}}+\textbf{\textrm{p}}_{\theta})$, respectively.
\begin{equation}
\mathcal{F}\left[\mathcal{F}^{-1}\left\{\tilde{S}_{u}(\textbf{\textrm{k}}-\textbf{\textrm{p}}_{\theta})\right\}\times e^{-i2\pi(\textbf{\textrm{p}}_{\theta}\cdot \textbf{\textrm{r}})}\right] = \tilde{S}_{s}(\textbf{\textrm{k}}-\textbf{\textrm{p}}_{\theta})
\label{eqShiftFreqComp1}
\end{equation}
\begin{equation}
\mathcal{F}\left[\mathcal{F}^{-1}\left\{\tilde{S}_{u}(\textbf{\textrm{k}}+\textbf{\textrm{p}}_{\theta})\right\}\times e^{+i2\pi(\textbf{\textrm{p}}_{\theta}\cdot \textbf{\textrm{r}})}\right] = \tilde{S}_{s}(\textbf{\textrm{k}}+\textbf{\textrm{p}}_{\theta})
\label{eqShiftFreqComp2}
\end{equation}
In the above equations, $\mathcal{F}$ and $\mathcal{F}^{-1}$ denotes Fourier transform and inverse Fourier transform, respectively.

\subsection{Phase matching}\label{secShiftPhaseError}
Note that from the consistency point of view,
\begin{equation}
\textrm{Arg}\left[\sum_{\textbf{k}}\tilde{S}_{u}(\textbf{\textrm{k}}) \tilde{S}^{*}_{s}(\textbf{k}-\textbf{\textrm{p}}_{\theta})\right]=0
\label{eqPhaseMatchCondition}
\end{equation}
i.e., the summation of phase mismatch between \textit{unshifted} central frequency component $\tilde{S}_{u}(\textbf{\textrm{k}})$, obtained through Eq.~(\ref{eqWienerFilter1}), and \textit{shifted} frequency component $\tilde{S}_{s}(\textbf{\textrm{k}}-\textbf{\textrm{p}}_{\theta})$, obtained through Eq.~(\ref{eqShiftFreqComp1}), over the frequencies $\textbf{\textrm{k}}$ where they overlap, must be zero. However, due to inaccuracies in the determination of phases $\phi_{1}$, $\phi_{2}$, $\phi_{3}$, condition~(\ref{eqPhaseMatchCondition}) inevitably gets violated. This calls for phase correction of shifted frequency components. The corrective phase $\phi_{c}$ is computed as
\begin{equation}
\phi_{c}=\textrm{Arg}\left[\sum_{\textbf{k}}\tilde{S}_{u}(\textbf{\textrm{k}}) \tilde{S}^{*}_{s}(\textbf{k}-\textbf{\textrm{p}}_{\theta})\right]
\label{eqShiftPhaseError}
\end{equation}
i.e., where summation is carried over the frequencies $\textbf{\textrm{k}}$ where $\tilde{S}_{u}(\textbf{\textrm{k}})$ and $\tilde{S}_{s}(\textbf{\textrm{k}}-\textbf{\textrm{p}}_{\theta})$ overlap. Subsequently, phase correction of shifted frequency components are effected as follows
\begin{equation}
\begin{array}{r}
\textrm{phase}\\
\textrm{corrected}
\end{array}
\left\{
\begin{array}{l}
\tilde{S}_{s}(\textbf{\textrm{k}}-\textbf{\textrm{p}}_{\theta}) = e^{-i\phi_{c}}\tilde{S}_{s}(\textbf{\textrm{k}}-\textbf{\textrm{p}}_{\theta}) \\ \tilde{S}_{s}(\textbf{\textrm{k}}+\textbf{\textrm{p}}_{\theta}) = e^{+i\phi_{c}}\tilde{S}_{s}(\textbf{\textrm{k}}+\textbf{\textrm{p}}_{\theta})
\end{array}
\right.
\label{eqShiftPhaseCorrection}
\end{equation}
where $\tilde{S}_{s}(\textbf{\textrm{k}}-\textbf{\textrm{p}}_{\theta})$ and $\tilde{S}_{s}(\textbf{\textrm{k}}+\textbf{\textrm{p}}_{\theta})$ on right hand side of Eq.~(\ref{eqShiftPhaseCorrection}) are those obtained by using Eqs.~(\ref{eqShiftFreqComp1}) and~(\ref{eqShiftFreqComp2}), respectively.

\subsection{Merging all frequency components using generalized Wiener filter}\label{secGenWF}

Using `approximate' generalized Wiener filter (Eq.~(\ref{EqWFgen3}) in Appendix), triplets -- $\tilde{S}_{u}(\textbf{\textrm{k}})$, $\tilde{S}_{s}(\textbf{\textrm{k}}-\textbf{\textrm{p}}_{\theta})$ and $\tilde{S}_{s}(\textbf{\textrm{k}}+\textbf{\textrm{p}}_{\theta})$ -- computed for each $\theta$ = $\theta_{1}$, $\theta_{2}$ and $\theta_{3}$ may be combined to obtain Fourier transform of SIM image.
\begin{eqnarray}
\tilde{D}_{\textrm{SIM}}(\textbf{\textrm{k}})&=&\sum^{\theta_{3}}_{\theta=\theta_{1}}\left[ \left(\frac{\mathcal{A}^{2}|\textbf{\textrm{k}}|^{-2\alpha}|\tilde{H}(\textbf{\textrm{k}})|^2/\Psi_{o,\theta}}{w+\Omega(\textbf{\textrm{k}})}\right)\tilde{S}_{u}(\textbf{\textrm{k}})\right. \nonumber \\
& &\hspace{-2.1cm}+\left(\frac{m^2\mathcal{A}^{2}|\textbf{\textrm{k}}-\textbf{\textrm{p}}_{\theta}|^{-2\alpha}|\tilde{H}(\textbf{\textrm{k}}+\textbf{\textrm{p}}_{\theta})|^2/\Psi_{p,\theta}}{w+ \Omega(\textbf{\textrm{k}})}\right)\tilde{S}_{s}(\textbf{\textrm{k}}-\textbf{\textrm{p}}_{\theta}) \nonumber \\
& &\hspace{-2.1cm}\left.+\left(\frac{m^2\mathcal{A}^{2}|\textbf{\textrm{k}}+\textbf{\textrm{p}}_{\theta}|^{-2\alpha}|\tilde{H}(\textbf{\textrm{k}}-\textbf{\textrm{p}}_{\theta})|^2/\Psi_{q,\theta}}{w+\Omega(\textbf{\textrm{k}})}\right)\tilde{S}_{s}(\textbf{\textrm{k}}+\textbf{\textrm{p}}_{\theta})\right] \nonumber \\
\label{EqWFgen3Do}
\end{eqnarray}
where
\begin{eqnarray}
\Omega(\textbf{\textrm{k}})&=& \sum^{\theta_{3}}_{\theta=\theta_{1}}\left[\frac{\mathcal{A}^{2}|\textbf{\textrm{k}}|^{-2\alpha}|\tilde{H}(\textbf{\textrm{k}})|^2}{\Psi_{o,\theta}} \right. \nonumber \\ &~&\hspace{0.7cm}+\frac{m^2\mathcal{A}^{2}|\textbf{\textrm{k}}-\textbf{\textrm{p}}_{\theta}|^{-2\alpha}|\tilde{H}(\textbf{\textrm{k}}+\textbf{\textrm{p}}_{\theta})|^2}{\Psi_{p,\theta}} \nonumber \\ &~&\hspace{0.7cm}\left.+\frac{m^2\mathcal{A}^{2}|\textbf{\textrm{k}}+\textbf{\textrm{p}}_{\theta}|^{-2\alpha}|\tilde{H}(\textbf{\textrm{k}}-\textbf{\textrm{p}}_{\theta})|^2}{\Psi_{q,\theta}}\right] \label{EqWFgen3DoA}
\end{eqnarray}
and $w$ is a constant, whose value is empirically adjusted so as to produce visibly optimum super-resolved image $D_{\textrm{SIM}}(\textbf{\textrm{r}})=F^{-1}[\tilde{D}_{\textrm{SIM}}(\textbf{\textrm{k}})]$.

In the above equation, $\tilde{H}(\textbf{\textrm{k}}\pm\textbf{\textrm{p}}_{\theta})$ is the system~OTF with its center shifted to frequency vector $\pm\textbf{\textrm{p}}_{\theta}$. However, when coordinates of $\pm\textbf{\textrm{p}}_{\theta}$ are not integral multiple of $1/N$ ($N\times N$ pixels being digital image size), it is important \emph{not} to use `Fourier shift theorem', as is employed in Eqs.~(\ref{eqShiftFreqComp1}) and~(\ref{eqShiftFreqComp2}), to shift system OTF $\tilde{H}(\textbf{\textrm{k}})$ to $\pm\textbf{\textrm{p}}_{\theta}$. The shifted $\tilde{H}(\textbf{\textrm{k}}\pm\textbf{\textrm{p}}_{\theta})$ obtained by such method is erroneous. This is a consequence of the discretization of frequencies which is inevitable while dealing with digital images and employing discrete Fourier transforms.

In the present work, the problem is resolved by observing that $\tilde{H}(\textbf{\textrm{k}}\pm\textbf{\textrm{p}}_{\theta})$ is not explicitly required, but only its power spectrum $|\tilde{H}(\textbf{\textrm{k}}\pm\textbf{\textrm{p}}_{\theta})|^{2}$ is required in Eq.~(\ref{EqWFgen3Do}). Consequently, an approximation $|\tilde{H}(\textbf{\textrm{k}}\pm\textbf{\textrm{p}}_{\theta})|^{2}$$\approx|\tilde{H}(\textbf{\textrm{k}}\pm\textbf{\textrm{p}}^{\textrm{rounded}}_{\theta})|^{2}$ is employed, where $\textbf{\textrm{p}}^{\textrm{rounded}}_{\theta}$ is frequency vector obtained by rounding of coordinates of  $\textbf{\textrm{p}}_{\theta}$ to nearest integral multiple of $1/N$. Now, since coordinates of $\textbf{\textrm{p}}^{\textrm{rounded}}_{\theta}$ are integral multiples of $1/N$, `Fourier shift theorem' is employed, as in Eqs.~(\ref{eqShiftFreqComp1}) and~(\ref{eqShiftFreqComp2}), to shift the system OTF $\tilde{H}(\textbf{\textrm{k}})$ to $\pm\textbf{\textrm{p}}^{\textrm{rounded}}_{\theta}$ and subsequently, the required power spectrum $|\tilde{H}(\textbf{\textrm{k}}\pm\textbf{\textrm{p}}_{\theta})|^{2}$ is computed.

\section{OpenSIM: an open source Matlab-based SIM-RA code}

The SIM-RA as described in section~\ref{secSIMalgorithm} (as well as the TIRF-SIM-RA described in section~\ref{secTIRFSIMalgorithm}), is coded into a series of Matlab-based script and function files. This complete set of files, collectively called as OpenSIM, is made freely available as a companion to this manuscript (see supplementary materials). This same set of files were used to obtain the results presented in section~\ref{secResults}.

\section{Results}\label{secResults}

\subsection{Simulation Results}

To evaluate the performance of SIM-RA, a synthetic image \texttt{testpat.1k.tiff}, which is essentially an image of Lenna appended by periodic patterns of varying fineness (Fig.~\ref{FigSimuResults}a), was used. The image \texttt{testpat.1k.tiff}, referred to as \textit{`test object'} henceforth in this manuscript, may be downloaded from USC-SIPI image database~\cite{sipi2005usc1}.

\subsubsection{Simulating raw SIM images}\label{secSimuSIMimages}

\begin{center}
\begin{figure}[h!]
\begin{pspicture}(0.0,0.2)(8.5,5.6)
\rput[lb](0.3,0.0){\scalebox{0.75}{\includegraphics[trim=118pt 501pt 177pt 135pt,clip]{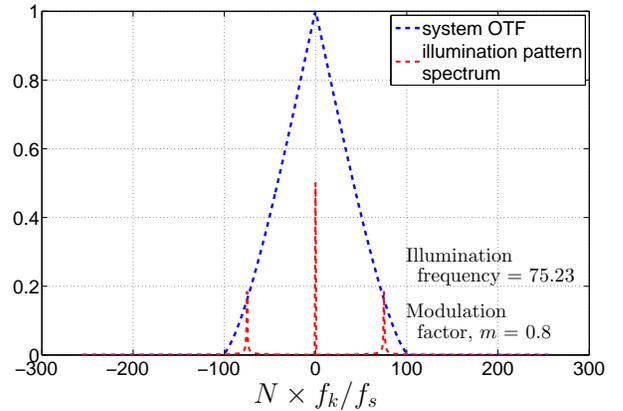}}}
\end{pspicture}
\caption{Optical transfer function $\tilde{H}(\textbf{\textrm{k}})$ and illumination pattern spectrum that was used for simulating raw SIM images. Note that the frequencies on the x-axis are indicated in terms of normalized frequency units: (actual frequency $f_{k}$)/(sampling frequency $f_{s}$). Since the size of the images used for generating raw SIM images is 512$\times$512 pixels, N=512.}
\label{FigOTFo}
\end{figure}
\end{center}

\begin{center}
\begin{figure*}[!t]
\begin{pspicture}(-1.0,0.0)(15,20.0)
\rput[lb](0.0,0.0){\scalebox{1.0}{\includegraphics[trim=89pt 206pt 84pt 72pt,clip]{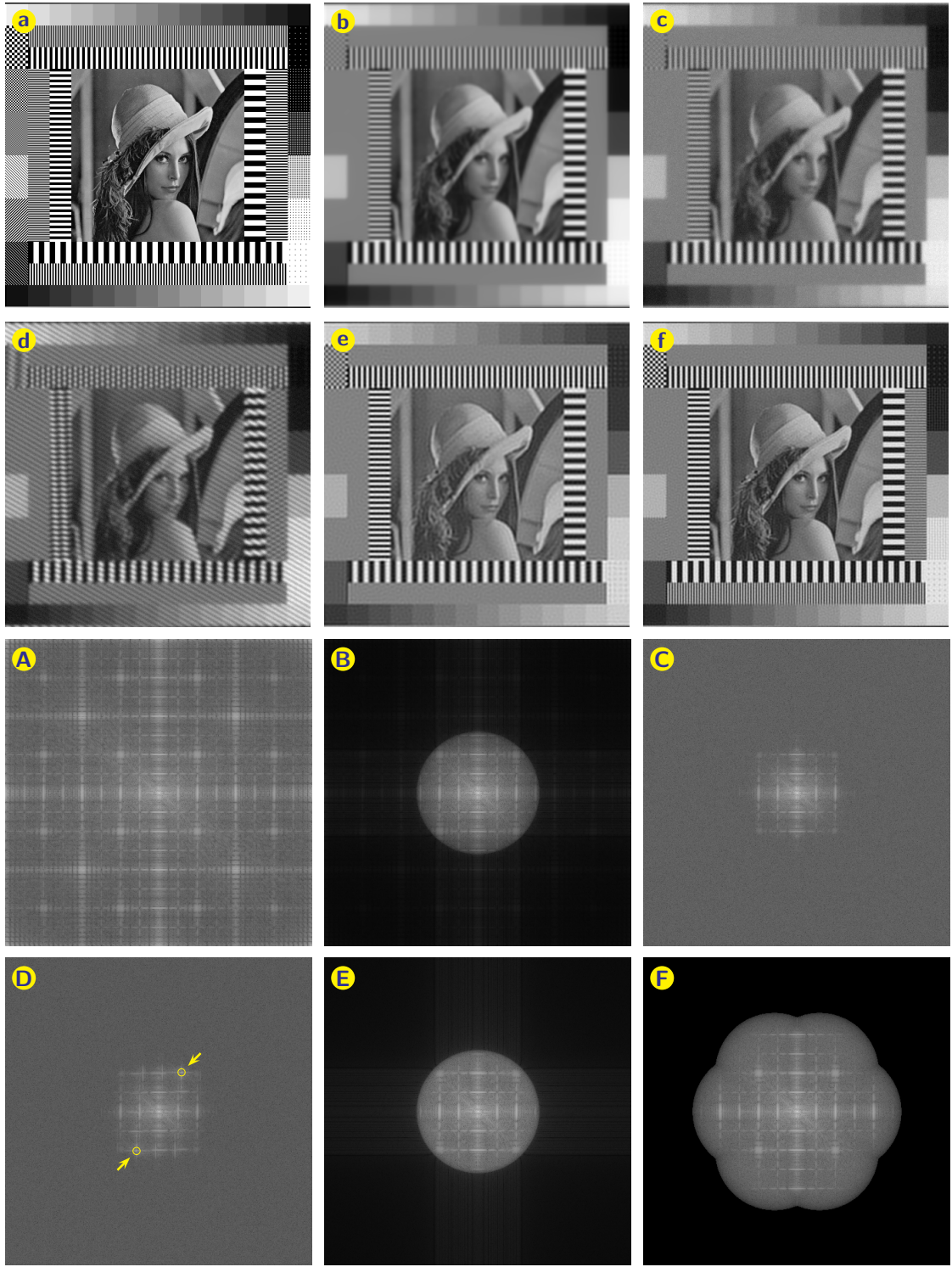}}}
\end{pspicture}
\caption{(a) Test object, (b) noise free image of Test object when imaged with optical system with OTF support as depicted in Fig.~\ref{FigOTFo}. (c) image (b) with 10\% (20db) Gaussian noise added. (d) simulated raw SIM image with sinusoidal illumination pattern. (e) Wiener-filtered estimate of noisy image (c). Reconstructed image using SIM-RA (Algorithm~\ref{AlgoSIMcode}). (A,B,...,F) Magnitude of Fourier transforms of images in (a,b,...,f).}
\label{FigSimuResults}
\end{figure*}
\end{center}

The optical transfer function $\tilde{H}(\textbf{\textrm{k}})$ of the optical imaging system is assumed to be circumsymmetric with a support width as depicted Fig.~\ref{FigOTFo}. Consequently, Fig.~\ref{FigSimuResults}b represents \textit{noise-free} image of test object as may be acquired by the optical system. However, it assumed that images acquired by the optical system are corrupted by an additive Gaussian noise of~10\% (i.e. 20db). Thus, Fig.~\ref{FigSimuResults}c depicts a more realistic (noise corrupted) image of the test object; the effect of noise addition is prominently observed in terms of increase in grey level beyond the OTF support in Fig.~\ref{FigSimuResults}C.

\vspace{-1.4cm}
Figure~\ref{FigOTFo} also depicts the spectrum of illumination pattern $I_{\theta,\phi}(\textbf{\textrm{r}})$ [see Eq.~(\ref{eqIllumination})] that was used for simulating raw SIM images. As discussed in sections~\ref{secShiftPhaseError},~\ref{secGenWF} and~\ref{secGWFdiscussion}, SIM reconstruction involves additional complications when illumination spatial frequency is not exact integral multiple $1/N$. To illustrate the effectiveness of developed SIM-RA, the illumination spatial frequency was deliberately chosen to be a fractional value, Fig.~\ref{FigOTFo}. The three orientations of the illumination pattern was chosen to be fixed at $\theta_{1}=0^{o}$, $\theta_{2}=60^{o}$ and $\theta_{3}=120^{o}$. However, phase shifts of the illumination pattern were assumed to be imprecise and simulated as $\phi_{1}=0^{o}+\phi_{e1}$, $\phi_{2}=120^{o}+\phi_{e2}$ and $\phi_{3}=240^{o}+\phi_{e3}$, where $\phi_{e1}$, $\phi_{e2}$ and $\phi_{e3}$ are random errors with uniform probability distribution function over the range $[-15^{o},15^{o}]$. Precise determination of phases $\phi_{1}$, $\phi_{2}$ and $\phi_{3}$ \textit{a posteriori} is incorporated within the SIM-RA. Figure~\ref{FigSimuResults}d depicts one of the nine raw SIM images thus simulated; the frequency peaks of illumination pattern are indicated by tiny circles in Fig.~\ref{FigSimuResults}D.

\subsubsection{Performance of SIM algorithm}\label{secPerformanceSIMalgorithm}

The SIM-RA (Algorithm~\ref{AlgoSIMcode}) described in section~\ref{secSIMalgorithm} was used to generate a super-resolution image (Figs.~\ref{FigSimuResults}f and~4F) of the test object using nine simulated raw SIM images (section~\ref{secSimuSIMimages}). Though the enhancement in resolution of the reconstructed image in comparison to widefield image (Fig.~\ref{FigSimuResults}c) is distinctively evident both in spatial and Fourier domain, it is more appropriate to study the resolution enhancement of the reconstructed image in comparison to the deconvolved (Wiener filtered) widefield image (Fig.~\ref{FigSimuResults}e). This is because a Wiener filtering operation is intrinsically present in SIM-RA (see section~\ref{secWienerFilter}).

\begin{center}
\begin{figure}[h!]
\begin{pspicture}(0.0,0.0)(8.4,5.4)
\rput[lb](0.1,0.0){\scalebox{0.75}{\includegraphics[trim=117pt 503pt 173pt 135pt,clip]{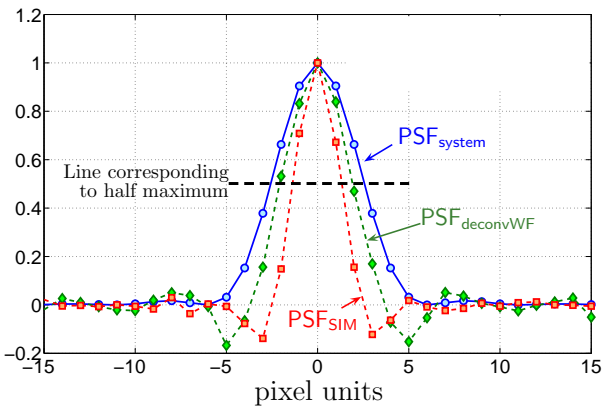}}}
\end{pspicture}
\caption{Numerically computed $\textrm{PSF}_{\textrm{deconvWF}}$ and $\textrm{PSF}_{\textrm{SIM}}$ together with imaging system PSF. Refer section~\ref{secPerformanceSIMalgorithm} for further details.}
\label{FigPSFe}
\end{figure}
\end{center}

Visual inspection of Figs.~\ref{FigSimuResults}e and~4f does reveal that the SIM reconstructed image is superior in resolution. Further, a quantitative study of resolution enhancement of the reconstructed image in comparison to the deconvolved widefield image was done by numerically determining the \emph{effective} point spread functions $\textrm{PSF}_{\textrm{deconvWF}}$ and $\textrm{PSF}_{\textrm{SIM}}$ with which the test image (Fig.~\ref{FigSimuResults}a) when convolved, respectively, produces deconvolved widefield image (Fig.~\ref{FigSimuResults}e) and SIM reconstructed image (Fig.~\ref{FigSimuResults}f). The determination of $\textrm{PSF}_{\textrm{deconvWF}}$ and $\textrm{PSF}_{\textrm{SIM}}$ is done employing linear algebra technique (see Appendix, section~\ref{secPSFdetermination}). The numerically computed $\textrm{PSF}_{\textrm{deconvWF}}$ and $\textrm{PSF}_{\textrm{SIM}}$ together with the imaging system PSF is depicted in Fig.~\ref{FigPSFe}. It may be noted that
\begin{eqnarray}
\textrm{FWHM}(\textrm{PSF}_{\textrm{deconvWF}})\hspace{-0.2cm}&\approx&\hspace{-0.2cm} 0.7\times\textrm{FWHM}(\textrm{PSF}_{\textrm{system}}) \label{eqPSFdeconv} \\
\textrm{FWHM}(\textrm{PSF}_{\textrm{SIM}})\hspace{-0.2cm}&\approx&\hspace{-0.2cm} 0.5\times\textrm{FWHM}(\textrm{PSF}_{\textrm{system}}) \label{eqPSFsim}
\end{eqnarray}
If $\lambda$ is the acquisition wavelength and $\textrm{NA}$ is the numerical aperture of objective of the optical system, then $\textrm{FWHM}(\textrm{PSF}_{\textrm{system}})=\lambda/(2\textrm{NA})$. Consequently, Eqs.~(\ref{eqPSFdeconv}) and~(\ref{eqPSFsim}) may be restated in a readily comprehensible form as follows
\begin{eqnarray}
\textrm{Resolution}_\textrm{(Deconv)} &\approx& \frac{1}{\sqrt{2}}\frac{\lambda}{2\textrm{NA}} \label{eqPResolutiondeconv} \\
\textrm{Resolution}_\textrm{(SIM)} &\approx& \frac{1}{2}\frac{\lambda}{2\textrm{NA}}, \label{eqResolutionSIM}
\end{eqnarray}
Note that $0.7\approx1/\sqrt{2}$; thus, Eq.~(\ref{eqPSFdeconv}) implies Eq.~(\ref{eqPResolutiondeconv}). Above relations between PSF widths suggest that SIM indeed improves the resolution by two-fold in spatial domain. Further, it may be noted that, the deconvolution process also improves the resolution by $\approx$1.4 fold~\cite{Wicker2014Resolving}.

Note that, theoretically, the maximum resolution enhancement that may be achieved through SIM-RA corresponds to $\textrm{FWHM}(\textrm{PSF}_{\textrm{SIM}})$ = 1/2$\times\textrm{FWHM}(\textrm{PSF}_{\textrm{deconvWF}})$. In the present case, this limit is not achieved because (1) illumination pattern frequency is not exactly at the edge of OTF cut-off frequency, but at $\approx$75\% of it (see Fig.~\ref{FigOTFo}), and (2) the noise level in the simulated SIM raw images is not zero but~10\%.

\vspace{-1.2cm}
\subsection{Experimental Results}\label{secSIMexpt}

\subsubsection{Acquisition of raw SIM images}\label{secExptSIMimages}
Raw SIM images of microtubules of~COS7 cells, stained first with anti-alpha tubulin antibody (Active Motif) and then with goat anti-mouse secondary antibody conjuncted with Alexa Fluor~488 (Life Technologies), were obtained using an optical system with the parameters: numerical aperture NA$=$1.49 (oil immersed), excitation wavelength $\lambda_{ex}$~=~488nm, emission wavelength $\lambda_{em}$~=~515nm, calibration~=~60nm/pixel. Figure~\ref{FigExptResults}a,d depicts one of the nine raw SIM images acquired; its fourier spectrum is represented in Figure~\ref{FigExptResults}A (illumination spatial frequency peaks are encircled and indicated by arrows).

\subsubsection{System OTF determination}\label{secExptOTF}
Several images of samples with sparsely distributed~100nm fluorescent microspheres were obtained. Intensity distribution corresponding to more than~100 microspheres were then super-imposed and averaged to obtain an approximation for system PSF; Fourier transform of this PSF provided an estimate of system OTF.

\subsubsection{Preprocessing of raw SIM images}\label{secExptSIMbackgroundSub}
The raw SIM images acquired experimentally were found to be corrupt with severe background fluorescence blur. Thus, raw SIM images were preprocessed in the following manner to make them suitable for SIM-RA:
\begin{enumerate}
  \item Intensity normalization: Raw SIM images were re-scaled so that all~9 images have identical global mean and standard deviation.
  \item Background fluorescence removal: The background of raw SIM fluorescent images were removed by employing in-built morphological operation function~\texttt{imopen} of Matlab~\cite{imopenMatlab}.
\end{enumerate}

\subsubsection{Reconstruction of high resolution image using SIM-RA}\label{secExptSIMreconstruction}

Direct application of SIM-RA on the processed SIM images typically produces reconstructed image with hexagonal pattern artifacts. This is due to presence of residual frequency peaks in vicinity of illumination spatial frequency in $\tilde{S}_{s}(\textbf{\textrm{k}}-\textbf{\textrm{p}}_{\theta})$ and $\tilde{S}_{s}(\textbf{\textrm{k}}+\textbf{\textrm{p}}_{\theta})$ for each of the three $\theta s$ resulting from inaccurate background fluorescence removal from raw SIM images; background removal techniques are heuristic in general.

\begin{center}
\begin{figure*}[h!]
\begin{pspicture}(-1.0,0.0)(15,15.0)
\rput[lb](0.0,0.0){\scalebox{1.0}{\includegraphics[trim=89pt 347pt 84pt 72pt,clip]{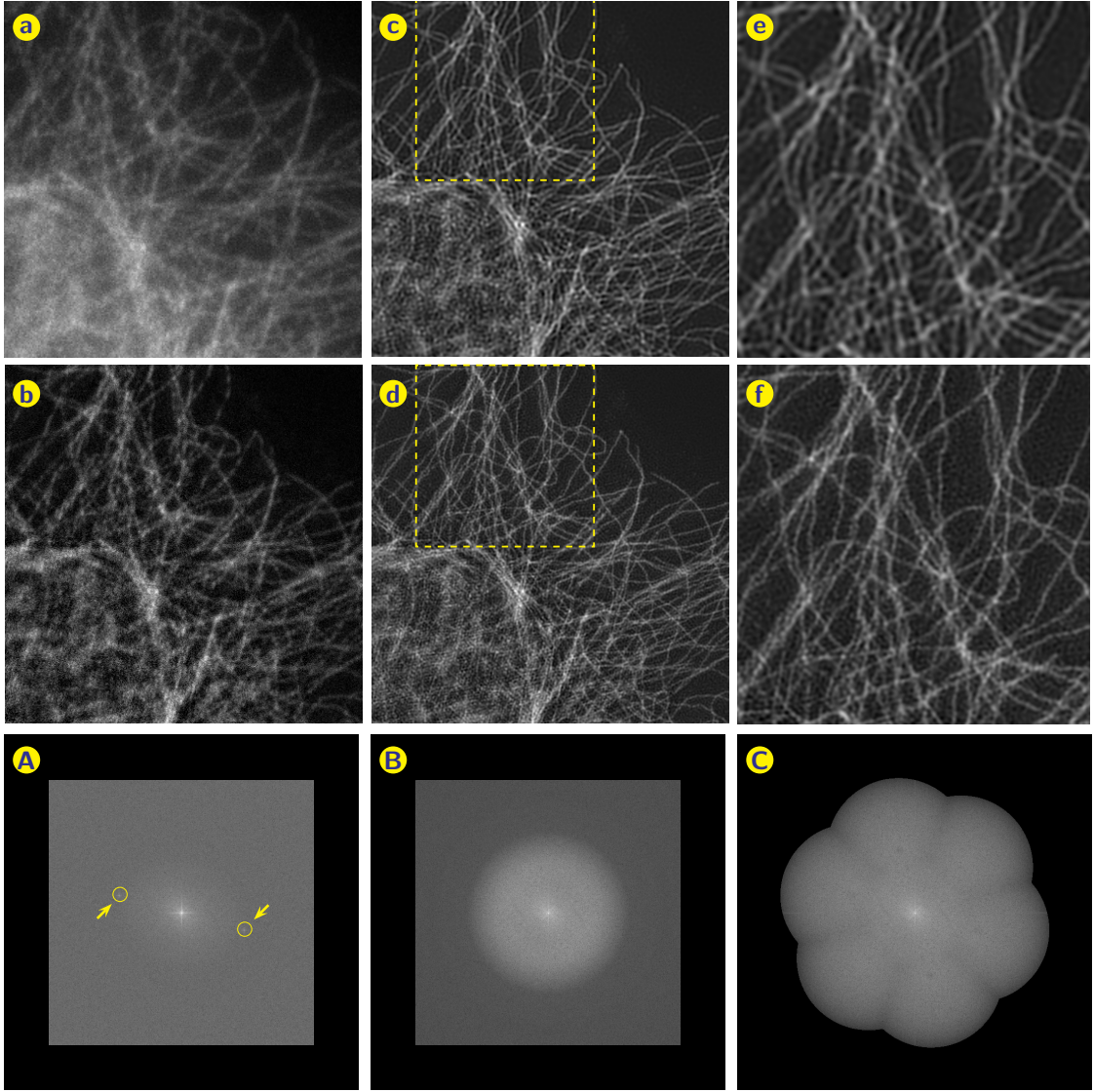}}}
\end{pspicture}
\caption{(a) Experimentally acquired raw SIM image with structured illumination; its Fourier spectrum depicting the illumination spatial frequency peaks (encircled and marked by arrows) is shown in (A). (b) Raw SIM image after background subtraction; its Fourier spectrum looks apparently similar to that shown in (A) and hence is not shown. (c) Image equivalent to Wiener filtered wide-field image (refer section~\ref{secExptSIMreconstruction}). (d) Reconstructed image $D_{\textrm{SIM}}(\textbf{\textrm{r}})$. Images~(e) \& (f) are magnified views of top region, indicated by dashed frame, of images~(c) \& (d) respectively. Fourier spectrum of images~a,c,d are indicated in A,B,C, respectively. [Specimen:~Immunofluorescent staining of microtubules of~COS7 cell.]}
\label{FigExptResults}
\end{figure*}
\end{center}

In order to suppress the effect of these residual peaks resulting from the inaccuracies in background fluorescence removal, the noisy estimates of $\tilde{S}(\textbf{\textrm{k}}-\textbf{\textrm{p}}_{\theta})\tilde{H}(\textbf{\textrm{k}})$ and $\tilde{S}(\textbf{\textrm{k}}+\textbf{\textrm{p}}_{\theta})\tilde{H}(\textbf{\textrm{k}})$ obtained at step~\ref{opeUnmix} of Algorithm~\ref{AlgoSIMcode} are multiplied by a heuristically designed notch filter
\begin{eqnarray}
F(\textbf{k}) &=& 1 - \exp(-a_{o}|\textbf{k}|^{\beta}) \label{eqNotchFilter}
\end{eqnarray}
where the parameters $a_{o}$ and $\beta$ are real constants, values of which are set by trial and error; in the present case  $a_{o}$ and $\beta$ were set to~0.05 and~1.2, respectively.  Rest of the steps of Algorithm~\ref{AlgoSIMcode} are implemented without change. The reconstructed image obtained by application of SIM-RA is depicted in Fig.~\ref{FigExptResults}d,f. Note that the effect of notch filter $F(\textbf{k})$ is barely visible in Fourier spectrum of reconstructed SIM image, Fig.~\ref{FigExptResults}C.

For comparison, an image equivalent to the deconvolved wide-field image, obtained by Wiener filtering the averaged central frequency component $\tilde{S}(\textbf{\textrm{k}})\tilde{H}(\textbf{\textrm{k}})$ determined in step~\ref{opeAAlpha} of Algorithm~\ref{AlgoSIMcode} is shown in Fig.~\ref{FigExptResults}c,e. The enhancement in resolution in the SIM-RA reconstructed image may be readily observed.


\vspace{-1.0cm}
\section{Discussion}

\subsection{Determination of illumination spatial frequency}

In published literature, illumination spatial frequency is tacitly assumed to be known \textit{a priori}~\cite{shroff2010lateral,wicker2013phase,wicker2013non}. The method used in the present work utilizes computation of auto-correlation of a single fourier image. This technique has its genesis in phase shift determination technique proposed by Wicker~\cite{wicker2013non}.

Note that determination of the illumination spatial frequency $\textbf{\textrm{p}}_{\theta}$ may directly be effected by iteratively optimizing the auto-correlation
\begin{equation}
\mathcal{C}_{1}=\sum_{\textbf{k}}\tilde{D}_{\theta,\phi}(\textbf{\textrm{k}})\tilde{D}^{*}_{\theta,\phi}(\textbf{k}+\textbf{\textrm{p}}_{\theta})
\label{eqFreqCC1a}
\end{equation}
too. However, multiplying $\tilde{D}_{\theta,\phi}(\textbf{\textrm{k}})$ with $\tilde{H}^{*}(\textbf{\textrm{k}})$ first, and then iteratively optimizing the auto-correlation as in Eq.~(\ref{eqFreqCC}) is superior due to the following two reasons: (1) it attenuates the effect of image white noise on the calculated value of $|\mathcal{C}_{1}|$, and (2) in case if system PSF is asymmetric, it neutralises any disturbing effect caused by it~\cite{wicker2013non}. It is remarked that reason~(2) renders no advantage in determination $\textbf{\textrm{p}}_{\theta}$ but reason~(1) does. This justifies the usage of Eq.~(\ref{eqFreqCC}) for computing auto-correlation, instead of Eq.~(\ref{eqFreqCC1a}), in section~\ref{secPtheta}.

\subsection{Determination of illumination phase shift}

Three different methods of illumination phase shift determination exists: (1)~phase-of-peak at illumination spatial frequency of patterned image computed in Fourier domain~\cite{shroff2010lateral}, (2)~iterative method based on minimizing the cross-correlation between the separated frequency components $\tilde{S}(\textbf{\textrm{k}}-\textbf{\textrm{p}}_{\theta})\tilde{H}(\textbf{\textrm{k}})$ and $\tilde{S}(\textbf{\textrm{k}}+\textbf{\textrm{p}}_{\theta})\tilde{H}(\textbf{\textrm{k}})$ (step~\ref{opeUnmix} of Algorithm~\ref{AlgoSIMcode})~\cite{wicker2013phase}, and (3)~phase determination based on autocorrelation of Fourier image~\cite{wicker2013non}.

The potential limitations of method~(1) is discussed in~\cite{shroff2010lateral,wicker2013phase}. Method~(3) is illustrated to be both efficient and accurate~\cite{wicker2013non}. The method essentially calls for computation of autocorrelation of
$\tilde{C}_{\theta,\phi}(\textbf{\textrm{k}})=\tilde{D}_{\theta,\phi}(\textbf{\textrm{k}}) \tilde{H}^{*}(\textbf{\textrm{k}})$ with its shifted variant $\tilde{C}_{\theta,\phi}(\textbf{k}+\textbf{\textrm{p}}_{\theta})$~:
\begin{equation}
\mathcal{C}_{3}=\sum_{\textbf{k}}\tilde{C}_{\theta,\phi}(\textbf{\textrm{k}})\tilde{C}^{*}_{\theta,\phi}(\textbf{k}+\textbf{\textrm{p}}_{\theta})
\label{eqPhaseCC}
\end{equation}
following which, an estimate of illumination phase is obtained as $\phi=-\textrm{Arg}(\mathcal{C}_{3})$.
However, raw SIM images acquired experimentally were found to be corrupted with background fluorescence blur (section~\ref{secExptSIMbackgroundSub}). Even though attempt is made to subtract it from the raw SIM images, its elimination is usually never perfect. Their presence introduces large errors in the frequency content of raw SIM images in a small neighborhood around zero frequency. This affects the accuracy of phase determination by both methods~(2) and~(3). Method~(1) which, unlike methods~(2) and~(3), uses only \emph{local} frequency information -- peak frequency at $\textbf{\textrm{p}}_{\theta}$ (or at $\textbf{\textrm{p}}^{\textrm{rounded}}_{\theta}$ (see section~\ref{secGenWF}) when the coordinates of $\textbf{\textrm{p}}_{\theta}$ are not integral multiple of $1/N$), which is far removed from the region where out-of-focus blur has erroneous effect -- potentially may produce a more accurate estimate of phase $\phi$ (but for the effect of additive Gaussian noise).

It is for these reasons that none of the three methods were employed in the present work. Instead phase~$\phi$ determination was effected in spatial domain as is described in section~\ref{secPhi}. The resulting estimates of phase~$\phi$ were accurate enough to produce an artifact-free image reconstruction even for experimental case, Fig.~\ref{FigExptResults}.

Note that for both simulated and experimental data, illumination spatial frequency lie within the system OTF support. The method described in section~\ref{secPhi} doesn't work when illumination spatial frequency lies beyond the system OTF support, as in TIRF-SIM. In such a case, iterative Method~(2) described above is be used. Section~\ref{secTIRFSIMalgorithm} describes a variant of Algorithm~\ref{AlgoSIMcode} that may be used to carry out reconstruction of SIM image when raw SIM images are obtained using a TIRF-SIM setup.

\subsection{Approximation for Generalized Wiener Filter}\label{secGWFdiscussion}

Direct usage of generalized Wiener filter, as presented in Eq.~(\ref{EqWFgen2}), for merging the separated components cannot be done in practice. This is because, if $\tilde{G}_{i}(\textbf{\textrm{k}})=\tilde{S}_{i}(\textbf{\textrm{k}})\tilde{H}_{i}(\textbf{\textrm{k}})$, and $\tilde{G}_{i}(\textbf{\textrm{k}}+\textbf{\textrm{p}}_{\theta})$, $\tilde{S}_{i}(\textbf{\textrm{k}}+\textbf{\textrm{p}}_{\theta})$ and $\tilde{H}_{i}(\textbf{\textrm{k}}+\textbf{\textrm{p}}_{\theta})$ are shifted forms of $\tilde{G}_{i}(\textbf{\textrm{k}})$, $\tilde{S}_{i}(\textbf{\textrm{k}})$ and $\tilde{H}_{i}(\textbf{\textrm{k}})$, respectively, to frequency center $\textbf{\textrm{p}}_{\theta}$, obtained using Fourier shift theorem, then $\tilde{G}_{i}(\textbf{\textrm{k}}+\textbf{\textrm{p}}_{\theta}) \neq\tilde{S}_{i}(\textbf{\textrm{k}}+\textbf{\textrm{p}}_{\theta})\tilde{H}_{i}(\textbf{\textrm{k}}+\textbf{\textrm{p}}_{\theta})$ when coordinates of $\textbf{\textrm{p}}_{\theta}$ are not integral multiple of $1/N$. Therefore, instances where equation of generalized Wiener filter, similar to Eq.~(\ref{EqWFgen2}), is presented in SIM literature, for example~\cite{shroff2010lateral,wicker2013phase}, are only meant to convey the general `conceptual idea' and \emph{not} the direct methodology for merging the separated frequency components.

Practical implementation of generalized Wiener filter necessarily requires approximations to be made and has an element of art to it; Gustafsson et al.~\cite{gustafsson2008three} have rightly used the adjective \emph{``somewhat non-intuitive''} to characterize the methodology of implementing generalized Wiener filter. The methodology presented in this manuscript, closely adheres to the approach suggested by Gustafsson et al.~\cite{gustafsson2008three}. Moreover, the present work makes the approach more explicit by presenting it in the form of a concrete algorithm.

\subsection{Apodization filter}

Usage of an apodization filter $\tilde{A}(\textbf{\textrm{k}})$ is advocated in some SIM-RA, refer~\cite{wicker2013phase} for instance. In these algorithms, $\tilde{D}_{\textrm{SIM}}(\textbf{\textrm{k}})$, as computed by Eq.~(\ref{EqWFgen3Do}), is multiplied by $\tilde{A}(\textbf{\textrm{k}})$, and then the SIM reconstructed image is computed as $D_{\textrm{SIM}}(\textbf{\textrm{r}})=$ $\mathcal{F}^{-1}\left[\tilde{D}_{\textrm{SIM}}(\textbf{\textrm{k}})\tilde{A}(\textbf{\textrm{k}})\right]$. An appropriately designed apodization filter helps avoid hard edges and ringing effect that may arise in reconstructed image $D_{\textrm{SIM}}(\textbf{\textrm{r}})$ otherwise.

In the present work, SIM reconstruction for the simulated case (Fig.~\ref{FigSimuResults}) is done without apodization filter, but for the experimental case (Fig.~\ref{FigExptResults})it was done using an apodization filter. OpenSIM package accompanied with this manuscript also possesses a subroutine to apodize filter $\tilde{D}_{\textrm{SIM}}(\textbf{\textrm{k}})$, as computed by Eq.~(\ref{EqWFgen3Do}), prior to reconstruction of SIM image. The apodization filter $\tilde{A}(\textbf{\textrm{k}})$ incorporated in OpenSIM is identical with that described in~\cite{wicker2013phase}.

\subsection{Customizing SIM-RA for Experimental data}

Pre-processing of raw SIM images as described in section~\ref{secExptSIMbackgroundSub} is necessary. Intensity normalization compensates for variations in overall intensities among the~9 raw SIM images. Background fluorescence removal reduces the overall blur in the reconstructed SIM image, which tends to mask the resolution enhancement. Since background fluorescence removal is realized heuristically, it is essential to employ a notch filter of the kind illustrated in Eq.~(\ref{eqNotchFilter}) to suppress residual peaks in lateral frequency components resulting from inaccuracies in background fluorescence removal. Such customization of SIM-RA is specific to the nature of experimental data and is done empirically.

\section{TIRF-SIM Algorithm}\label{secTIRFSIMalgorithm}

In a TIRF-SIM setup, the illumination spatial frequency typically lies beyond the system OTF support. This section describes a variant of Algorithm~\ref{AlgoSIMcode} that may be used to carry out reconstruction of SIM image when the raw SIM images are obtained using a TIRF-SIM setup.

\begin{center}
\begin{algorithm*}[!t]
    \SetKwInOut{Input}{Input}
    \SetKwInOut{Output}{Output}

    \underline{function TIRF-SIM} \;
    \Input{System OTF $\tilde{H}(\textbf{\textrm{k}})$ and nine raw SIM images $D_{\theta,\phi}(\textbf{\textrm{r}})$, corresponding to  $\theta=\theta_{1},\theta_{2},\theta_{3}$ and $\phi=\phi_{1},\phi_{2},\phi_{3}$  }
    \Output{Reconstructed SIM image $D_{SIM}(\textbf{\textrm{r}})$}
    \For{$\left\{D_{\theta,\phi}(\textbf{\textrm{r}})|~\theta=\theta_{1},\theta_{2},\theta_{3}\right\}$}
    {
        Estimate phases $\phi'_{2}$ and $\phi'_{3}$ (section~\ref{secWickerPhase})\; \label{opePhasesTIRF}

        Obtain noisy estimates of $\tilde{S}(\textbf{\textrm{k}})\tilde{H}(\textbf{\textrm{k}})$, $\tilde{S}_{\circ}(\textbf{\textrm{k}}-\textbf{\textrm{p}}_{\theta})\tilde{H}(\textbf{\textrm{k}})$ and $\tilde{S}_{\circ}(\textbf{\textrm{k}}+\textbf{\textrm{p}}_{\theta})\tilde{H}(\textbf{\textrm{k}})$ using Eq.~(\ref{eqInvApproxTIRF}) (do this by setting $m=1$ in matrix $\textbf{\textrm{M}}_{\circ}$ [Eq.~(\ref{eqImageMatrixTIRF})]; Adjustment for $m\neq1$ is effected later in steps~\ref{opeMfTIRF} and~\ref{opeWFcomponentsTIRF} below) \; \label{opeUnmixTIRF}

        Estimate illumination spatial frequency $\textbf{\textrm{p}}_{\theta}$ (section~\ref{secWickerPtheta})\;

    }
    Average the three noisy estimates of $\tilde{S}(\textbf{\textrm{k}})\tilde{H}(\textbf{\textrm{k}})$ (one for each $\theta=\theta_{1},\theta_{2},\theta_{3}$) and use it to estimate parameters $\mathcal{A}$ and $\alpha$ characterizing object power spectrum (section~\ref{secAAlpha}) \; 

    \For{$\theta=\theta_{1},\theta_{2},\theta_{3}$}
    {
        Determine modulation factor $m$ (following the line of reasoning provided in section~\ref{secMFdetermination})\; \label{opeMfTIRF}

        Use Wiener Filter to obtain noise-filtered and ungraded estimates $\tilde{S}_{u}(\textbf{\textrm{k}})$, $\tilde{S}_{u}(\textbf{\textrm{k}}-\textbf{\textrm{p}}_{\theta})$ and $\tilde{S}_{u}(\textbf{\textrm{k}}+\textbf{\textrm{p}}_{\theta})$ of noisy $\tilde{S}(\textbf{\textrm{k}})\tilde{H}(\textbf{\textrm{k}})$, $\tilde{S}_{\circ}(\textbf{\textrm{k}}-\textbf{\textrm{p}}_{\theta})\tilde{H}(\textbf{\textrm{k}})$ and $\tilde{S}_{\circ}(\textbf{\textrm{k}}+\textbf{\textrm{p}}_{\theta})\tilde{H}(\textbf{\textrm{k}})$, respectively (section~\ref{secWienerFilter}; Note: This requires replacing $\tilde{S}(\textbf{\textrm{k}}-\textbf{\textrm{p}}_{\theta})\tilde{H}(\textbf{\textrm{k}})$ and $\tilde{S}(\textbf{\textrm{k}}+\textbf{\textrm{p}}_{\theta})\tilde{H}(\textbf{\textrm{k}})$ in Eqs.~(\ref{eqWienerFilter2}) and~(\ref{eqWienerFilter3}) with $\tilde{S}_{\circ}(\textbf{\textrm{k}}-\textbf{\textrm{p}}_{\theta})\tilde{H}(\textbf{\textrm{k}})$ and $\tilde{S}_{\circ}(\textbf{\textrm{k}}+\textbf{\textrm{p}}_{\theta})\tilde{H}(\textbf{\textrm{k}})$, respectively.)\; \label{opeWFcomponentsTIRF}

        Shift $\tilde{S}_{u}(\textbf{\textrm{k}}-\textbf{\textrm{p}}_{\theta})$ and $\tilde{S}_{u}(\textbf{\textrm{k}}+\textbf{\textrm{p}}_{\theta})$ components to their respective true positions in frequency domain (section~\ref{secShiftFreqComp}). Let $\tilde{S}_{s}(\textbf{\textrm{k}}-\textbf{\textrm{p}}_{\theta})$ and $\tilde{S}_{s}(\textbf{\textrm{k}}+\textbf{\textrm{p}}_{\theta})$, respectively, denote components $\tilde{S}_{u}(\textbf{\textrm{k}}-\textbf{\textrm{p}}_{\theta})$ and $\tilde{S}_{u}(\textbf{\textrm{k}}+\textbf{\textrm{p}}_{\theta})$ shifted to their true positions\; \label{opeTIRFpreShiftCorrection}

        `Phase match' the shifted components $\tilde{S}_{s}(\textbf{\textrm{k}}-\textbf{\textrm{p}}_{\theta})$ and $\tilde{S}_{s}(\textbf{\textrm{k}}+\textbf{\textrm{p}}_{\theta})$ with respect to the unshifted component $\tilde{S}_{u}(\textbf{\textrm{k}})$ (section~\ref{secShiftPhaseError}) \; \label{opeTIRFShiftCorrection}
    }
    Merge all nine frequency frequency components (three components $\tilde{S}_{u}(\textbf{\textrm{k}})$, $\tilde{S}_{s}(\textbf{\textrm{k}}-\textbf{\textrm{p}}_{\theta})$ and $\tilde{S}_{s}(\textbf{\textrm{k}}+\textbf{\textrm{p}}_{\theta})$ for each of the three $\theta s$) into one $\tilde{D}_{\textrm{SIM}}(\textbf{\textrm{k}})$ using generalized Wiener-Filter (section~\ref{secGenWF}) \; 

    Compute inverse FT of $\tilde{D}_{SIM}(\textbf{\textrm{k}})$ to obtain reconstructed SIM image $D_{\textrm{SIM}}(\textbf{\textrm{r}})$ \;
    return $D_{\textrm{SIM}}(\textbf{\textrm{r}})$ \;
    \caption{TIRF-SIM-RA}
    \label{AlgoTIRFSIMcode}
\end{algorithm*}
\end{center}

\subsection{Formulation}

Defining
\begin{eqnarray}
\tilde{S}_{\circ}(\textbf{\textrm{k}}-\textbf{\textrm{p}}_{\theta})&=&e^{-i\phi_{1}}\tilde{S}(\textbf{\textrm{k}}-\textbf{\textrm{p}}_{\theta}) \label{eqSrightTIRF} \\
\tilde{S}_{\circ}(\textbf{\textrm{k}}+\textbf{\textrm{p}}_{\theta})&=&e^{+i\phi_{1}}\tilde{S}(\textbf{\textrm{k}}+\textbf{\textrm{p}}_{\theta}) \label{eqSleftTIRF} \\
\phi'_{2} &=& \phi_{2} - \phi_{1} \label{eqPhase2TIRF} \\
\phi'_{3} &=& \phi_{3} - \phi_{1} \label{eqPhase3TIRF}
\end{eqnarray}
Eq.~(\ref{eqImageMatrix}) may be rewritten as
\begin{equation}
\left[\hspace{-0.1cm} \begin{array}{c}
\tilde{D}_{\theta,\phi_{1}}(\textbf{\textrm{k}}) \\
\tilde{D}_{\theta,\phi_{2}}(\textbf{\textrm{k}}) \\
\tilde{D}_{\theta,\phi_{3}}(\textbf{\textrm{k}}) \end{array} \hspace{-0.1cm}\right]= \frac{I_{o}}{2}\textbf{\textrm{M}}_{\circ}\hspace{-0.1cm} \left[\hspace{-0.1cm} \begin{array}{c}
\tilde{S}(\textbf{\textrm{k}})\tilde{H}(\textbf{\textrm{k}}) \\
\tilde{S}_{\circ}(\textbf{\textrm{k}}-\textbf{\textrm{p}}_{\theta})\tilde{H}(\textbf{\textrm{k}}) \\
\tilde{S}_{\circ}(\textbf{\textrm{k}}+\textbf{\textrm{p}}_{\theta})\tilde{H}(\textbf{\textrm{k}}) \end{array} \hspace{-0.1cm}\right] + \left[\hspace{-0.1cm} \begin{array}{c}
\tilde{N}_{\theta,\phi_{1}}(\textbf{\textrm{k}}) \\
\tilde{N}_{\theta,\phi_{2}}(\textbf{\textrm{k}}) \\
\tilde{N}_{\theta,\phi_{3}}(\textbf{\textrm{k}}) \end{array} \hspace{-0.1cm}\right] \label{eqSuperposedMatTIRF}
\end{equation}
\begin{equation}
\textrm{where}~~\textbf{\textrm{M}}_{\circ}=\left[ \begin{array}{lll}
1 & -\frac{m}{2} & -\frac{m}{2} \\
1 & -\frac{m}{2}e^{-i\phi'_{2}} & -\frac{m}{2}e^{+i\phi'_{2}} \\
1 & -\frac{m}{2}e^{-i\phi'_{3}} & -\frac{m}{2}e^{+i\phi'_{3}} \end{array} \right] \label{eqImageMatrixTIRF}
\end{equation}
Setting $I_{o}/2=1$, analogous to Eq.~(\ref{eqInvApprox}), we have
\begin{equation}
\begin{array}{r}
\textrm{noisy }\\
\textrm{estimate}\\
\textrm{of} \end{array}
\left[ \begin{array}{c}
\tilde{S}(\textbf{\textrm{k}})\tilde{H}(\textbf{\textrm{k}}) \\
\tilde{S}_{\circ}(\textbf{\textrm{k}}-\textbf{\textrm{p}}_{\theta})\tilde{H}(\textbf{\textrm{k}}) \\
\tilde{S}_{\circ}(\textbf{\textrm{k}}+\textbf{\textrm{p}}_{\theta})\tilde{H}(\textbf{\textrm{k}}) \end{array} \right] = \textbf{\textrm{M}}^{-1}_{\circ}\left[ \begin{array}{c}
\tilde{D}_{\theta,\phi_{1}}(\textbf{\textrm{k}}) \\
\tilde{D}_{\theta,\phi_{2}}(\textbf{\textrm{k}}) \\
\tilde{D}_{\theta,\phi_{3}}(\textbf{\textrm{k}}) \end{array} \right]
\label{eqInvApproxTIRF}
\end{equation}

\subsection{Determination of phases $\phi'_{2}$ and $\phi'_{3}$}\label{secWickerPhase}
Let $\psi_{2}$ and $\psi_{3}$ be \emph{approximate} estimates of $\phi'_{2}$ and $\phi'_{3}$, respectively. Then, approximation to $\textbf{\textrm{M}}_{\circ}$ would be
\begin{equation*}
\textbf{\textrm{M}}_{e} = \left[ \begin{array}{ccc}
1 & -\frac{1}{2} & -\frac{1}{2} \\
1 & -\frac{1}{2}e^{-i\psi_{2}} & -\frac{1}{2}e^{+i\psi_{2}} \\
1 & -\frac{1}{2}e^{-i\psi_{3}} & -\frac{1}{2}e^{+i\psi_{3}} \end{array} \right]
\end{equation*}
and $\textbf{\textrm{M}}^{-1}_{e}=$
\begin{equation}
\frac{1}{\Delta_{e}}\hspace{-0.1cm}\left[\hspace{-0.1cm} \begin{array}{ccc}
e^{i\left(\psi_{2}-\psi_{3}\right)}-e^{i\left(\psi_{3}-\psi_{2}\right)} & e^{i\psi_{3}}-e^{-i\psi_{3}} & e^{i\psi_{2}}-e^{-i\psi_{2}} \\
2\left(e^{i\psi_{3}}-e^{i\psi_{2}}\right) & 2\left(1-e^{i\psi_{3}}\right) & 2\left(e^{i\psi_{2}}-1\right) \\
2\left(e^{-i\psi_{2}}-e^{-i\psi_{3}}\right) & 2\left(e^{-i\psi_{3}}-1\right) & 2\left(1-e^{-i\psi_{2}}\right)
\end{array}\hspace{-0.1cm} \right]\label{eqnMinvLongE}
\end{equation}
where
\begin{eqnarray*}
\Delta_{e}\hspace{-0.3cm}&=&\hspace{-0.3cm}\left[e^{i\phi_{2}} -e^{-i\psi_{2}} -e^{i\psi_{3}} +e^{-i\psi_{3}}+e^{i\left(\psi_{3}-\psi_{2}\right)}-e^{i\left(\psi_{2}-\psi_{3}\right)}\right] 
\end{eqnarray*}
Let $\tilde{S}_{a}(\textbf{\textrm{k}})\tilde{H}(\textbf{\textrm{k}})$, $\tilde{S}_{a}(\textbf{\textrm{k}}-\textbf{\textrm{p}}_{\theta})\tilde{H}(\textbf{\textrm{k}})$ and $\tilde{S}_{a}(\textbf{\textrm{k}}-\textbf{\textrm{p}}_{\theta})\tilde{H}(\textbf{\textrm{k}})$ be noisy estimates of $\tilde{S}(\textbf{\textrm{k}})\tilde{H}(\textbf{\textrm{k}})$, $\tilde{S}_{\circ}(\textbf{\textrm{k}}-\textbf{\textrm{p}}_{\theta})\tilde{H}(\textbf{\textrm{k}})$ and $\tilde{S}_{\circ}(\textbf{\textrm{k}}-\textbf{\textrm{p}}_{\theta})\tilde{H}(\textbf{\textrm{k}})$, respectively, when $\psi_{2}$ and $\psi_{3}$ are used as \emph{approximate} estimates $\phi'_{2}$ and $\phi'_{3}$. Then, from Eq.~(\ref{eqInvApproxTIRF}) we have
\begin{eqnarray}
\left[ \begin{array}{c}
\tilde{S}_{a}(\textbf{\textrm{k}})\tilde{H}(\textbf{\textrm{k}}) \\
\tilde{S}_{a}(\textbf{\textrm{k}}-\textbf{\textrm{p}}_{\theta})\tilde{H}(\textbf{\textrm{k}}) \\
\tilde{S}_{a}(\textbf{\textrm{k}}+\textbf{\textrm{p}}_{\theta})\tilde{H}(\textbf{\textrm{k}}) \end{array} \right] \hspace{-0.2cm}&=&\hspace{-0.2cm} \textbf{\textrm{M}}^{-1}_{e}\left[ \begin{array}{c}
\tilde{D}_{\theta,\phi_{1}}(\textbf{\textrm{k}}) \\
\tilde{D}_{\theta,\phi_{2}}(\textbf{\textrm{k}}) \\
\tilde{D}_{\theta,\phi_{3}}(\textbf{\textrm{k}}) \end{array} \right] \label{EqRawSideLobs} \\ \hspace{-0.2cm}&=&\hspace{-0.2cm}\textbf{\textrm{M}}^{-1}_{e}\textbf{\textrm{M}}_{\circ}\left[ \begin{array}{c}
\tilde{S}(\textbf{\textrm{k}})\tilde{H}(\textbf{\textrm{k}}) \\
\tilde{S}_{\circ}(\textbf{\textrm{k}}-\textbf{\textrm{p}}_{\theta})\tilde{H}(\textbf{\textrm{k}}) \\
\tilde{S}_{\circ}(\textbf{\textrm{k}}+\textbf{\textrm{p}}_{\theta})\tilde{H}(\textbf{\textrm{k}}) \end{array} \right] \nonumber
\end{eqnarray}
where Eq.~(\ref{eqSuperposedMatTIRF}) is used in the last equality; noise term is ignored for compactness. The above equation may be solved to obtain
\begin{equation}
\left[\hspace{-0.1cm} \begin{array}{c}
\tilde{S}_{a}(\textbf{\textrm{k}}-\textbf{\textrm{p}}_{\theta})\tilde{H}(\textbf{\textrm{k}}) \\
\tilde{S}_{a}(\textbf{\textrm{k}}+\textbf{\textrm{p}}_{\theta})\tilde{H}(\textbf{\textrm{k}}) \end{array} \hspace{-0.1cm}\right] = \frac{m}{\Delta_{e}}\left[\hspace{-0.1cm} \begin{array}{cc}
a_{22} & a_{23} \\
a_{32} & a_{33} \end{array} \hspace{-0.1cm}\right] \left[\hspace{-0.1cm} \begin{array}{c}
\tilde{S}_{\circ}(\textbf{\textrm{k}}-\textbf{\textrm{p}}_{\theta})\tilde{H}(\textbf{\textrm{k}}) \\
\tilde{S}_{\circ}(\textbf{\textrm{k}}+\textbf{\textrm{p}}_{\theta})\tilde{H}(\textbf{\textrm{k}}) \end{array} \hspace{-0.1cm}\right] \label{EqApproxSideLobs}
\end{equation}
where
\begin{eqnarray*}
a_{22} \hspace{-0.2cm}&=&\hspace{-0.2cm} e^{i\psi_{2}} -e^{-i\phi'_{2}} -e^{i\psi_{3}} +e^{-i\phi'_{3}}+e^{i\left(\psi_{3}-\phi'_{2}\right)}-e^{i\left(\psi_{2}-\phi'_{3}\right)} \\
a_{23} \hspace{-0.2cm}&=&\hspace{-0.2cm} e^{i\psi_{2}} -e^{i\phi'_{2}} -e^{i\psi_{3}} +e^{i\phi'_{3}} +e^{i\left(\psi_{3}+\phi'_{2}\right)} -e^{i\left(\psi_{2}+\phi'_{3}\right)} \nonumber \\
a_{32} \hspace{-0.2cm}&=&\hspace{-0.2cm} e^{-i\phi'_{2}} -e^{-i\psi_{2}} -e^{-i\phi'_{3}}  +e^{-i\psi_{3}} +e^{-i\left(\phi'_{3}+\psi_{2}\right)} \nonumber \\
\hspace{-0.2cm}&~&\hspace{-0.2cm} \hspace{5.4cm}-e^{-i\left(\phi'_{2}+\psi_{3}\right)} \nonumber \\
a_{33} \hspace{-0.2cm}&=&\hspace{-0.2cm} e^{i\phi'_{2}} -e^{-i\psi_{2}} -e^{i\phi'_{3}} +e^{-i\psi_{3}} +e^{i\left(\phi'_{3}-\psi_{2}\right)}-e^{i\left(\phi'_{2}-\psi_{3}\right)} \nonumber \end{eqnarray*}
Note that $a_{23}=0$ and $a_{32}=0$ only when $\psi_{2}=\phi'_{2}$ and $\psi_{3}=\phi'_{3}$. For this condition, referring to Eq.~(\ref{EqApproxSideLobs}), $\tilde{S}_{a}(\textbf{\textrm{k}}-\textbf{\textrm{p}}_{\theta})\tilde{H}(\textbf{\textrm{k}})= m\tilde{S}_{\circ}(\textbf{\textrm{k}}-\textbf{\textrm{p}}_{\theta})\tilde{H}(\textbf{\textrm{k}})$ and has no contribution from $\tilde{S}_{\circ}(\textbf{\textrm{k}}+\textbf{\textrm{p}}_{\theta})\tilde{H}(\textbf{\textrm{k}})$; similarly, $\tilde{S}_{a}(\textbf{\textrm{k}}+\textbf{\textrm{p}}_{\theta})\tilde{H}(\textbf{\textrm{k}})= m\tilde{S}_{\circ}(\textbf{\textrm{k}}+\textbf{\textrm{p}}_{\theta})\tilde{H}(\textbf{\textrm{k}})$ and has no contribution from $\tilde{S}_{\circ}(\textbf{\textrm{k}}-\textbf{\textrm{p}}_{\theta})\tilde{H}(\textbf{\textrm{k}})$.  This ensures that cross-correlation between $\tilde{S}_{a}(\textbf{\textrm{k}}-\textbf{\textrm{p}}_{\theta})\tilde{H}(\textbf{\textrm{k}})$ and $\tilde{S}_{a}(\textbf{\textrm{k}}+\textbf{\textrm{p}}_{\theta})\tilde{H}(\textbf{\textrm{k}})$ is minimum when $\psi_{2}=\phi'_{2}$ and $\psi_{3}=\phi'_{3}$. This fact is used to determine the unknown phases $\phi'_{2}$ and $\phi'_{3}$~\cite{wicker2013phase}.

Specifically, beginning with arbitrary initial guesses for unknown phases $\phi'_{2}$ and $\phi'_{3}$, say $\psi_{2}$ and $\psi_{3}$, respectively, $\tilde{S}_{a}(\textbf{\textrm{k}}-\textbf{\textrm{p}}_{\theta})\tilde{H}(\textbf{\textrm{k}})$ and $\tilde{S}_{a}(\textbf{\textrm{k}}+\textbf{\textrm{p}}_{\theta})\tilde{H}(\textbf{\textrm{k}})$ are computed using Eq.~(\ref{EqRawSideLobs}). Following this, we compute the cross-correlation
\begin{equation}
\mathcal{C}_{4} = \sum_{\textbf{\textrm{k}}}w(\textbf{\textrm{k}}) \left(\tilde{S}_{a}(\textbf{\textrm{k}}-\textbf{\textrm{p}}_{\theta})\tilde{H}(\textbf{\textrm{k}})\right) \left(\tilde{S}_{a}(\textbf{\textrm{k}}+\textbf{\textrm{p}}_{\theta})\tilde{H}(\textbf{\textrm{k}})\right)^{*} \label{eqCCwTIRF}
\end{equation}
where $w(\textbf{\textrm{k}})$ is a weighing function to minimize the effect of noise in $\tilde{S}_{a}(\textbf{\textrm{k}}-\textbf{\textrm{p}}_{\theta})\tilde{H}(\textbf{\textrm{k}})$ and $\tilde{S}_{a}(\textbf{\textrm{k}}+\textbf{\textrm{p}}_{\theta})\tilde{H}(\textbf{\textrm{k}})$ on the computed value of $\mathcal{C}_{4}$. Subsequently, the values of $\psi_{2}$ and $\psi_{3}$ are iteratively optimized to produce a minima for $|\mathcal{C}_{4}|$. Values of $\psi_{2}$ and $\psi_{3}$ for the optimum condition are taken to be the estimates of unknown phases $\phi'_{2}$ and $\phi'_{3}$, respectively. For Gaussian noise, optimal weighing function is shown to be $w(\textbf{\textrm{k}})=\tilde{H}(\textbf{\textrm{k}})\tilde{H}^{*}(\textbf{\textrm{k}})$~\cite{wicker2013phase}.

\subsection{Determination of illumination spatial frequency $\textbf{\textrm{p}}_{\theta}$}\label{secWickerPtheta}

An estimate of illumination spatial frequency $\textbf{\textrm{p}}_{\theta}$ is obtained by iteratively optimizing $\textbf{\textrm{p}}_{e}$ to obtain a maxima for the normalized cross-power spectrum~\cite{foroosh2002extension}
\begin{equation}
\mathcal{C}_{5} = \left|\frac{\sum_{\textbf{\textrm{k}}} \tilde{S}_{\textrm{central}}(\textbf{\textrm{k}}) \tilde{S}^{*}_{\textrm{side}}(\textbf{\textrm{k}}+\textbf{\textrm{p}}_{e})}{\sum_{\textbf{\textrm{k}}} \tilde{S}_{\textrm{side}}(\textbf{\textrm{k}}+\textbf{\textrm{p}}_{e}) \tilde{S}^{*}_{\textrm{side}}(\textbf{\textrm{k}}+\textbf{\textrm{p}}_{e})}\right| \label{eqCCfreqTIRF}
\end{equation}
The noisy estimates of $\tilde{S}_{\circ}(\textbf{\textrm{k}})\tilde{H}(\textbf{\textrm{k}})$ and $\tilde{S}_{\circ}(\textbf{\textrm{k}}-\textbf{\textrm{p}}_{\theta})\tilde{H}(\textbf{\textrm{k}})$ obtained in step~\ref{opeUnmixTIRF} of Algorithm~\ref{AlgoTIRFSIMcode} are used to define the following intermediate terms which are required for evaluation of $\mathcal{C}_{5}$.
\begin{eqnarray*}
\tilde{S}_{\textrm{central}}(\textbf{\textrm{k}}) &=& \left[\tilde{S}(\textbf{\textrm{k}})\tilde{H}(\textbf{\textrm{k}})\right] \tilde{H}^{*}(\textbf{\textrm{k}}) \label{eqCCfreqTIRF1} \\
\tilde{S}_{\textrm{side}}(\textbf{\textrm{k}}) &=&  \left[\tilde{S}_{\circ}(\textbf{\textrm{k}}-\textbf{\textrm{p}}_{\theta})\tilde{H}(\textbf{\textrm{k}})\right] \tilde{H}^{*}(\textbf{\textrm{k}}) \label{eqCCfreqTIRF0} \\
\tilde{S}_{\textrm{side}}(\textbf{\textrm{k}}+\textbf{\textrm{p}}_{e}) &=&  \mathcal{F}\left[\left\{\mathcal{F}^{-1}\tilde{S}_{\textrm{side}}(\textbf{\textrm{k}})\right\}\times e^{-i2\pi(\textbf{\textrm{p}}_{e}\cdot \textbf{\textrm{r}})}\right]
\label{eqCCfreqTIRF2}
\end{eqnarray*}
The fact that the overlapping frequency contents of $\tilde{S}(\textbf{\textrm{k}})\tilde{H}(\textbf{\textrm{k}})$ and $\tilde{S}_{\circ}(\textbf{\textrm{k}}-\textbf{\textrm{p}}_{\theta})\tilde{H}(\textbf{\textrm{k}})$, shifted to its correct location, are correlated, ensures that $\mathcal{C}_{5}$ achieves its maxima when $\textbf{\textrm{p}}_{e}=\textbf{\textrm{p}}_{\theta}$.

\subsection{Remarks on Algorithm~\ref{AlgoTIRFSIMcode}}
(1) Algorithm~\ref{AlgoSIMcode} estimates illumination spatial frequency first and then estimates illumination phase. In Algorithm~\ref{AlgoTIRFSIMcode}, this order is reversed.

(2) Unlike Algorithm~\ref{AlgoSIMcode}, Algorithm~\ref{AlgoTIRFSIMcode} never estimates true illumination phases $\phi_{1}$, $\phi_{2}$ and $\phi_{3}$. Algorithm~\ref{AlgoTIRFSIMcode} estimates \textit{relative} phases $\phi'_{2}$ $(=\phi_{2}-\phi_{1})$ and $\phi'_{3}$ $(=\phi_{3}-\phi_{1})$ in step~\ref{opePhasesTIRF}. The information of phase $\phi_{1}$ is absorbed within estimates of $\tilde{S}_{\circ}(\textbf{\textrm{k}}-\textbf{\textrm{p}}_{\theta})$ and $\tilde{S}_{\circ}(\textbf{\textrm{k}}+\textbf{\textrm{p}}_{\theta})$ computed in step~\ref{opeUnmixTIRF}. During subsequent steps of Algorithm~\ref{AlgoTIRFSIMcode}, the information of phase $\phi_{1}$ propagates to $\tilde{S}_{s}(\textbf{\textrm{k}}-\textbf{\textrm{p}}_{\theta})$ and $\tilde{S}_{s}(\textbf{\textrm{k}}+\textbf{\textrm{p}}_{\theta})$ computed in step~\ref{opeTIRFpreShiftCorrection}. Eventually, the `phase matching' effected in step~\ref{opeTIRFShiftCorrection} relieves $\tilde{S}_{s}(\textbf{\textrm{k}}-\textbf{\textrm{p}}_{\theta})$ and $\tilde{S}_{s}(\textbf{\textrm{k}}+\textbf{\textrm{p}}_{\theta})$ of the information of phase $\phi_{1}$ contained within them. Thus, unlike the `phase matching' step of Algorithm~\ref{AlgoSIMcode}, which serves to correct the minor imprecisions in the estimation of \textit{absolute} illumination phases $\phi_{1}$, $\phi_{2}$ and $\phi_{3}$, the `phase matching' step of Algorithm~\ref{AlgoTIRFSIMcode} compensates for the fact that all prior computations were carried out using the estimates of \textit{relative} illumination phases  $\phi'_{2}$ and $\phi'_{3}$.

(3) Algorithm~\ref{AlgoTIRFSIMcode} may also be used to carry out SIM image reconstruction when illumination spatial frequency lies within the OTF support, as for the cases depicted in Figs.~\ref{FigSimuResults} and~\ref{FigExptResults}. The reconstructed SIM images, for both simulated and experimental cases described in section~\ref{secResults}, obtained using Algorithm~\ref{AlgoTIRFSIMcode} were found to be visually similar to that depicted in Figs.~\ref{FigSimuResults} and~\ref{FigExptResults}.

\begin{center}
\begin{figure*}[t!]
\begin{pspicture}(0.0,0.0)(14,8.0)
\rput[lb](2.2,0.0){\scalebox{1.0}{\includegraphics[trim=64pt 547pt 211pt 81pt,clip]{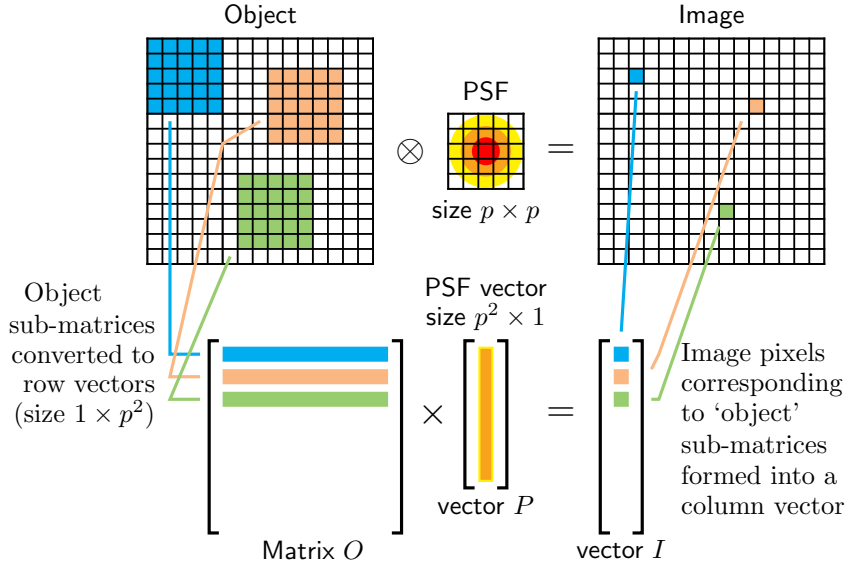}}}
\end{pspicture}
\caption{\textit{Top Panel:} Each `image' pixel which is more than half PSF width away from edge is obtained as a result of weighted-average of a sub-matrix of `object' with PSF.\textit{ Bottom panel:} By appropriately rearranging the elements of object's sub-matrix into a row vector and that of PSF's into a column vector ($P$), this weighted-average operation may be expressed as product of these row and column vectors. Consequently, a matrix equation: $O\times P=I$ may be formed. This equation may be used to estimate PSF, in case when it is unknown but `object' and `image' are known.}
\label{FigPSFdetermination}
\end{figure*}
\end{center}
\vspace{-1.0cm}

\section{Conclusion}

In the family of super-resolution microscopy, SIM plays an important role because it brings insight from the view point of frequency domain, whereas other techniques primarily do so from the spatial domain. It requires the interference between the illumination spatial frequency and the sample frequency, so that the differential frequency falling within the OTF support of the microscopic objective can be detected. Such modulation with illumination pattern is done at several different angles and phases. Consequently, a post-processing to bring the relevant frequency components back to their original coordinates in frequency domain, is necessary. In this work, we collect all the details -- such as the determination of the illumination spatial frequency, phase shifts, objective power spectrum, modulation factor, phase shift error, translocation and merging of various frequency components -- into one coherent piece to build a SIM-RA. Following this, we demonstrate efficacy of SIM-RA for simulated SIM images. Specifically, we studied the resolution enhancement that may be achieved through direct deconvolution (Wiener filtering) and SIM reconstruction by computing effective PSFs. Previously, the $2\times$ resolution enhancement of SIM has always been assessed in frequency domain, but never been illustrated in spatial domain with PSF width computation. Here we show that, deconvolution such as Wiener filtering can provide $\approx1.4\times$ resolution enhancement, whereas SIM provides $\approx2\times$ resolution enhancement in term of reduction in PSF width. Further, we demonstrate the efficacy of SIM-RA for experimental SIM data, by high-resolution image reconstruction of the microtubules in COS7 cell.

The SIM-RA as described in the manuscript, is coded into a series of Matlab-based script and function files. This complete set of files, collectively called as OpenSIM, is freely made available as a companion to this manuscript. This is to enable the users to better understand the algorithm, and to manipulate the parameters interactively, so that the errors/artifacts of SIM reconstruction can be minimized. It is hoped that readers from both instrumentation background and biological background may find this paper resourceful.

In this work, we have restricted our discussion on 2D SIM. Based on the same principle, 3D SIM can also be done in a similar manner, which is an ongoing project of the authors.


%
\appendices
\section{Wiener Filtering multiple but non identical images of same object}

Let $G_{i}(\textbf{\textrm{r}})$ be a set of $n$ images capturing different views $S_{i}(\textbf{\textrm{r}})$ of same object $S(\textbf{\textrm{r}})$, such that
\begin{equation}
G_{i}(\textbf{\textrm{r}})=S_{i}(\textbf{\textrm{r}})\otimes H_{i}(\textbf{\textrm{r}})+N_{i}(\textbf{\textrm{r}}) \hspace{0.5cm}\left\{i=1,2,\ldots,n\right\} \label{eqWFgen0}
\end{equation}
where $H_{i}(\textbf{\textrm{r}})$ is system PSF and $N_{i}(\textbf{\textrm{r}})$ is additive noise in each image. In fourier domain, Eq.~(\ref{eqWFgen0}) is given by
\begin{equation}
\tilde{G}_{i}(\textbf{\textrm{k}})=\tilde{S}_{i}(\textbf{\textrm{k}}) \tilde{H}_{i}(\textbf{\textrm{k}})+\tilde{N}_{i}(\textbf{\textrm{k}}) \hspace{0.5cm}\left\{i=1,2,\ldots,n\right\} \label{eqWFgen1}
\end{equation}
In such situation, an estimate $S_{a}(\textbf{\textrm{r}})$ of ungraded object $S(\textbf{\textrm{r}})$ is provided by generalized Wiener Filter~\cite{yaroslavsky1994deconvolution}
\begin{equation}
\tilde{S}_{a}(\textbf{\textrm{k}})=\sum^{n}_{i=1} \left[\frac{\Phi_{i}(\textbf{\textrm{k}})|\tilde{H}_{i}(\textbf{\textrm{k}})|^2/\Psi_{i}}{1+\sum^{n}_{i=1}\Phi_{i}(\textbf{\textrm{k}})|\tilde{H}_{i}(\textbf{\textrm{k}})|^2/\Psi_{i}}\right]\frac{\tilde{G}_{i}(\textbf{\textrm{k}})}{\tilde{H}_{i}(\textbf{\textrm{k}})}
\label{EqWFgen2}
\end{equation}
where\\
\indent$|\tilde{H}_{i}(\textbf{\textrm{k}})|^{2}=\tilde{H}_{i}(\textbf{\textrm{k}})\tilde{H}^{\ast}_{i}(\textbf{\textrm{k}})=$ power spectrum of $i^{\textrm{th}}$ OTF \\
\indent$\Psi_{i}=$ average noise power in $i^{\textrm{th}}$ image\\
\indent$\Phi_{i}(\textbf{\textrm{k}})=|\overline{\tilde{S}}_{i}(\textbf{\textrm{k}})|^{2}=$ mean power spectrum of $i^{\textrm{th}}$ ungraded image.

\subsection{Approximation to Generalized Wiener Filter}\label{secModifiedWF}

Conventional Wiener Filter estimate of $\tilde{S}_{i}(\textbf{\textrm{k}})$ is given by
\begin{equation}
\tilde{S}_{o,i}(\textbf{\textrm{k}})=\left[\frac{|\tilde{H}_{i}(\textbf{\textrm{k}})|^{2}}{|\tilde{H}_{i}(\textbf{\textrm{k}})|^{2} +\Psi_{i}/\Phi_{i}(\textbf{\textrm{k}})}\right] \frac{\tilde{G}_{i}(\textbf{\textrm{k}})}{\tilde{H}_{i}(\textbf{\textrm{k}})}
\label{EqWFgen4}
\end{equation}
Approximating $\tilde{G}_{i}(\textbf{\textrm{k}})/\tilde{H}_{i}(\textbf{\textrm{k}})\approx\tilde{S}_{o,i}(\textbf{\textrm{k}})$ (note: this approximation becomes equality when noise power $\Psi_{i}=0$),
Eq.~(\ref{EqWFgen2}) may be modified as
\begin{equation}
\tilde{S}_{a}(\textbf{\textrm{k}})=\sum^{n}_{i=1} \left[\frac{\Phi_{i}(\textbf{\textrm{k}})|\tilde{H}_{i}(\textbf{\textrm{k}})|^2/\Psi_{i}}{w+\sum^{n}_{i=1}\Phi_{i}(\textbf{\textrm{k}})|\tilde{H}_{i}(\textbf{\textrm{k}})|^2/\Psi_{i}}\right] \tilde{S}_{o,i}(\textbf{\textrm{k}}) \label{EqWFgen3}
\end{equation}
Note that the additive constant~1 in denominator of Eq.~(\ref{EqWFgen2}) is replaced by a constant parameter $w$ whose value needs to be empirically set. Since, ${S}_{o,i}(\textbf{\textrm{k}})$ is already noise-filtered, Eq.~(\ref{EqWFgen4}), suitable value of parameter $w$ may be empirically searched for in the range $\{0<w\leq1\}$.

\section{Effective PSF determination}\label{secPSFdetermination}

It is possible to express the convolution operation between `object' and PSF resulting in `image' formation in the form of a matrix multiplication equation, see Fig.~\ref{FigPSFdetermination}.

Consequently, when both `object' and `image' are known, it is possible to estimate the unknown PSF. Begin with a wise estimate of the size of PSF, say $p\times p$. Then using the `object' and `image', construct a matrix $O$ and a vector $I$, as illustrated in Fig.~\ref{FigPSFdetermination}, both with number of rows $r>p^2$. Note that, first, vector $I$ is formed by randomly selecting $r>p^2$ image pixels (all these pixels must be more than half PSF width away from image edge) and arranging them into a column vector. Corresponding to each image pixel that is selected for constructing $I$, its corresponding sub-matrix in object is located; elements of this sub-matrix are rearranged to form row of matrix $O$. Since, $O\times P=I$, least square solution for PSF vector is given by $P=(O^{T}O)^{-1}O^{T}I$. Subsequently, elements of vector $P$ may be rearranged to obtain 2-dimension PSF of size $p\times p$.

For determination of both $\textrm{PSF}_{\textrm{deconvWF}}$ and $\textrm{PSF}_{\textrm{SIM}}$ (see section~\ref{secPerformanceSIMalgorithm}), PSF size was assumed to be $40\times40$ pixels. While solving for PSFs, matrix $O$ and vector $I$ were constructed with number of rows $r=7\times40\times40$. By constructing different pairs $O$ and $I$, PSF vector $P$ was solved for,~100 times; mean of all these estimated $P$s was then used to reconstruct the PSF under determination, eventually.

\section*{Acknowledgment}
We thank Prof. Shaoqiang Tang and Prof. K Satish for helpful discussions. This work was supported by the National Instrument Development Special Program (2013YQ03065102),  and the National Natural Science Foundation of China (61178076, 61475010, 31327901).



\bibliographystyle{IEEEtran}
%

\bibliography{References1}

%

\begin{IEEEbiography}[{\includegraphics[width=1in,height=1.25in,clip,keepaspectratio]{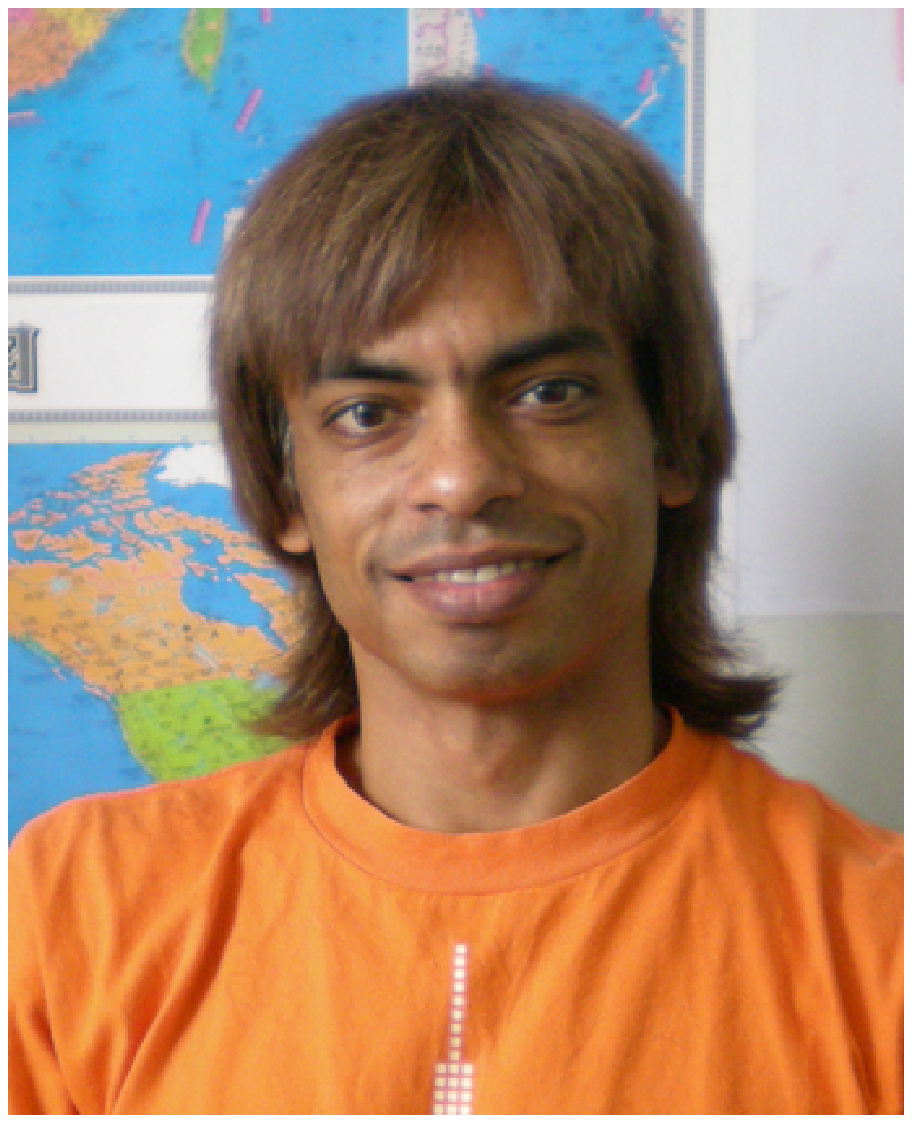}}]{Amit Lal}
Amit Lal received his Ph.D. from Dept. of Aerospace Engineering, Indian Institute of Science, Bangalore, India. Following this, he worked as a Research Assistant for three years in the field of Respiratory Neuroscience at Hyogo College of Medicine, Nishinomiya, Japan. He is currently a doctorate student in Dept. of Biomedical Engineering, College of Engineering, Peking University.
\end{IEEEbiography}

\begin{IEEEbiography}[{\includegraphics[width=1in,height=1.25in,clip,keepaspectratio]{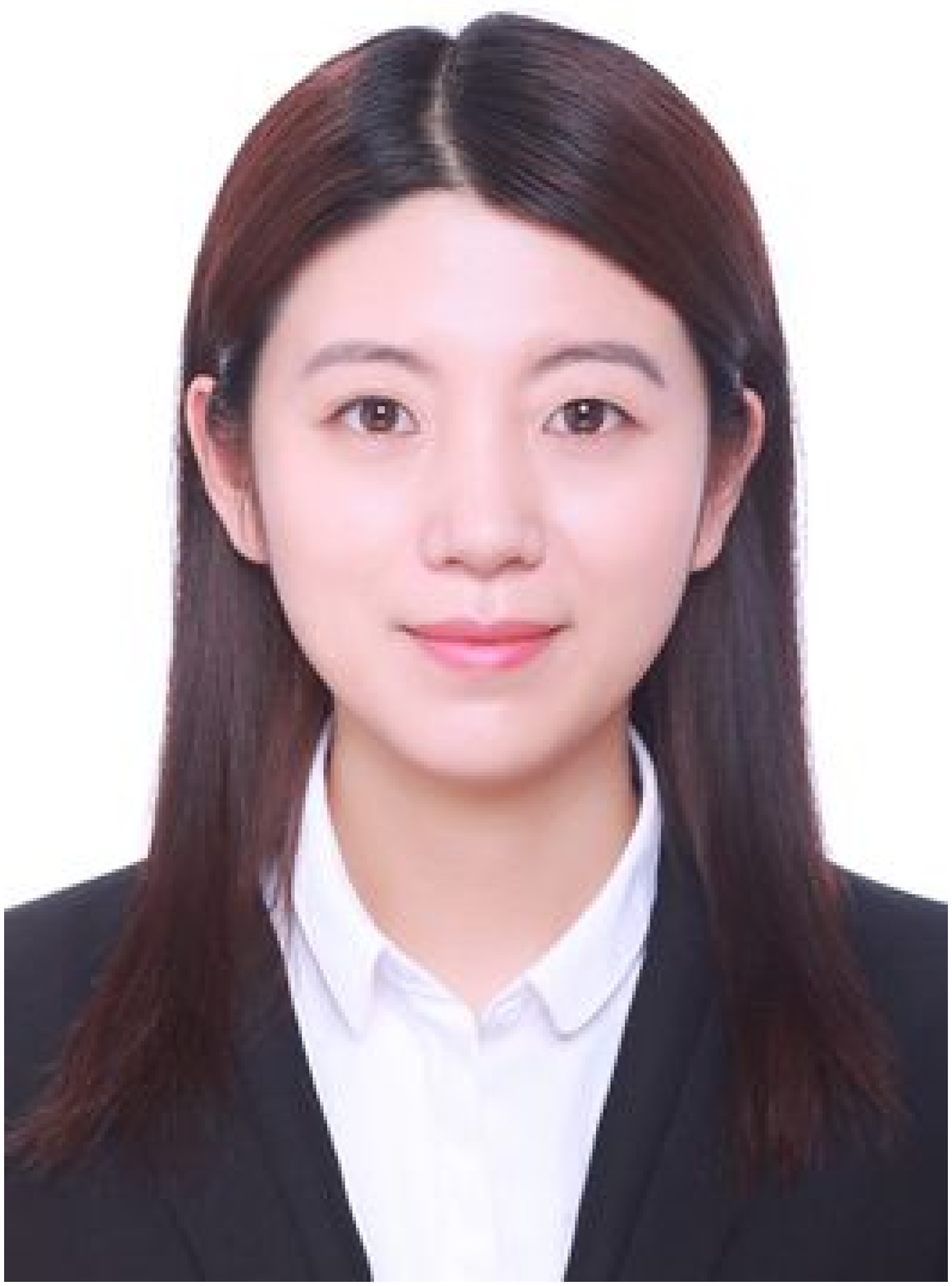}}]{Chunyan Shan}
Chunyan Shan obtained her Ph. D. from School of Life Sciences, Peking University, Beijing, China. Afterwards, she did her postdoctoral study in Core Facilities of Life Sciences, Peking University. Since 2015, she works in Core Facilities of Life Sciences, Peking University as an Engineer.

Dr. Chunyan Shan’s current research interests focus on application of super-resolution microscopy in cell biology, for instance, live-cell super-resolution imaging, ultrastructure analysis using SIM, STED and STORM.

\end{IEEEbiography}


\begin{IEEEbiography}[{\includegraphics[width=1in,height=1.25in,clip,keepaspectratio]{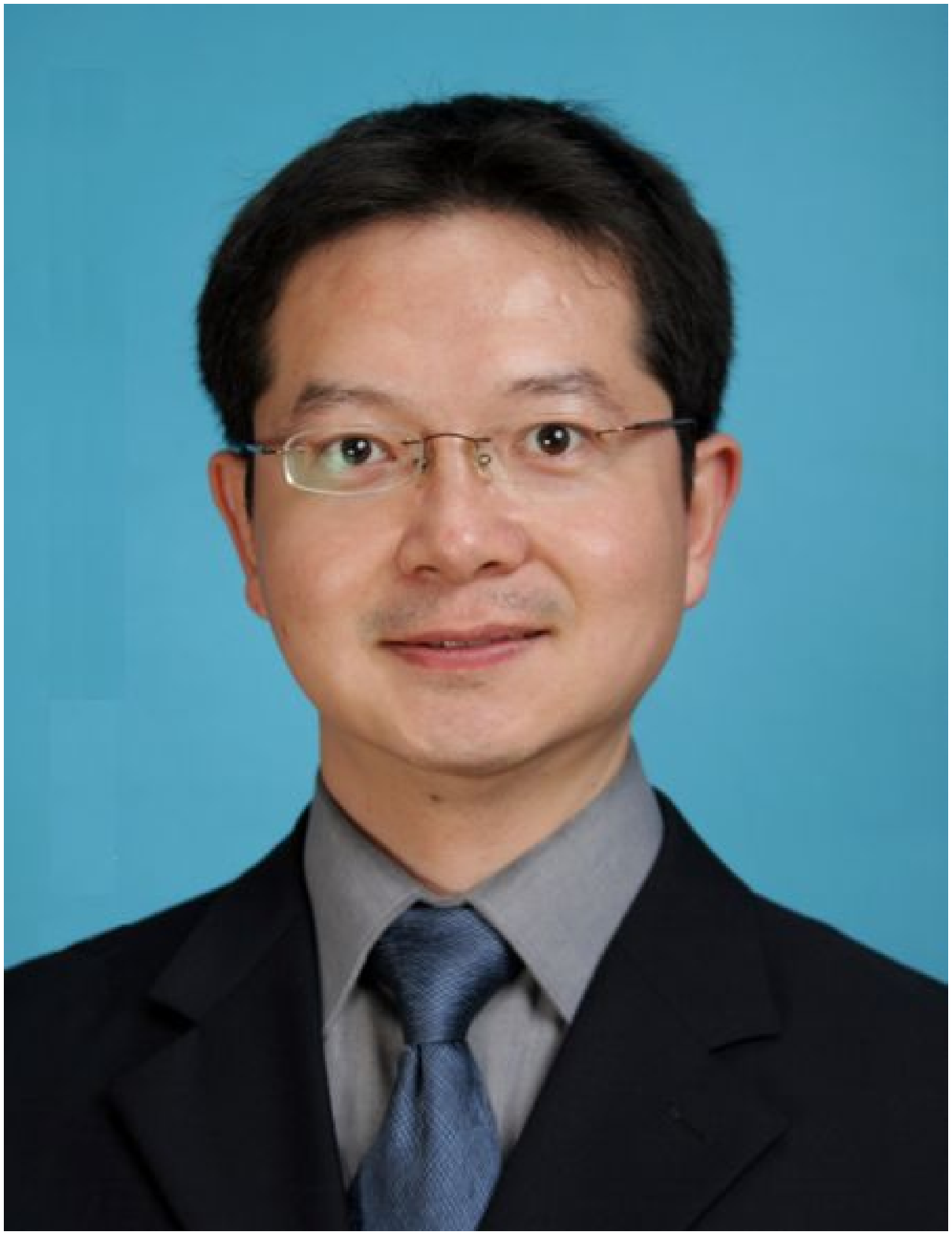}}]{Peng Xi}
Dr. Peng Xi obtained his Ph. D. from Shanghai Institute of Optics and Fine Mechanics, Chinese Academy of Sciences, Shanghai, China. He then worked as a postdoctoral Research Associate in three universities: Hong Kong University of Science and Technology, Purdue University, and Michigan State University. From~2008 to~2009 he worked in the Department of Biomedical Engineering, Shanghai Jiao Tong University as an Associate Professor. He is now working in Dept. of Biomedical Engineering, College of Engineering, Peking University as an Associate Professor since 2009.

Dr. Peng Xi's current research interests are focused on research and development of optical nanoscopy, as well as confocal and multiphoton microscopy. Dr. Peng Xi has published over 40 scientific papers in peer-reviewed journals such as Nature Photonics, ACS Nano, Scientific Reports, Optics Letters, Optics Express, etc., and received 7 issued patents, including 2 US patents.

Dr. Peng Xi is on the editorial board of several SCI-indexed journals: Scientific Reports, Micron, Microscopy Research and Techniques, and Chinese Optics Letters. He has been invited to give several invited talks in international conferences hosted by IEEE, OSA, and SPIE.

\end{IEEEbiography}



\end{document}